\documentclass{article}

\usepackage[utf8]{inputenc}
\usepackage[T1]{fontenc}
\usepackage{main}
\usepackage{microtype}
\usepackage{graphicx}
\usepackage{times}
\usepackage{inconsolata}
\usepackage{amsmath}
\usepackage{amssymb}
\usepackage{enumitem}
\usepackage{booktabs}
\usepackage{textcomp}
\usepackage{xspace}
\usepackage{float}
\usepackage{multirow}
\usepackage{fancyhdr}
\usepackage{wrapfig}
\usepackage{soul}
\usepackage[dvipsnames]{xcolor}

\definecolor{mydarkblue}{rgb}{0,0.08,0.45}
\usepackage[colorlinks=true,linkcolor=mydarkblue,citecolor=mydarkblue,filecolor=mydarkblue,urlcolor=mydarkblue]{hyperref}
\setlist[itemize,1]{leftmargin=10pt}

\usepackage{tikz}
\usepackage{xcolor}
\usetikzlibrary{arrows.meta}
\definecolor{TaxBlueBg}{HTML}{EDF4FA}
\definecolor{TaxBlueHead}{HTML}{D7E8F7}
\definecolor{TaxBlueLine}{HTML}{5C8DB8}
\definecolor{TaxGreenBg}{HTML}{EEF7F1}
\definecolor{TaxGreenHead}{HTML}{D8ECD9}
\definecolor{TaxGreenLine}{HTML}{6CA47A}
\definecolor{TaxRedBg}{HTML}{F9ECEC}
\definecolor{TaxRedHead}{HTML}{F2D6D6}
\definecolor{TaxRedLine}{HTML}{C78383}
\definecolor{TaxSlate}{HTML}{667085}
\usepackage{soul}
\usepackage{amssymb}
\usepackage{amsmath}
\usepackage{xspace}
\usepackage{float}
\usepackage{booktabs}
\usepackage{colortbl}
\usepackage{subcaption}
\usepackage{caption}
\usepackage{bbm}
\usepackage{adjustbox}
\usepackage{makecell}
\usepackage[framemethod=TikZ]{mdframed}

\definecolor{mypositive}{RGB}{0, 128, 0}
\definecolor{mynegative}{RGB}{220, 20, 60}
\definecolor{mypositive}{RGB}{24, 103, 173}
\definecolor{mynegative}{RGB}{255, 128, 0}

\definecolor{customblue}{HTML}{1F77B4} 
\definecolor{customorange}{HTML}{FF7F0E}
\definecolor{customgreen}{HTML}{2CA02C}
\definecolor{no}{HTML}{BA8E23}

\definecolor{darkgreen}{RGB}{0,100,0}
\definecolor{darkred}{RGB}{139,0,0}

\newcommand{\deltaScore}[1]{%
\begingroup
\ifdim #1pt > 0pt
\textcolor{darkred}{\scriptsize$^{#1}$}%
\else
\textcolor{darkgreen}{\scriptsize$^{#1}$}%
\fi
\endgroup
}

\usepackage[skins,breakable]{tcolorbox}

\definecolor{neutralgray}{HTML}{6B7280}
\definecolor{neutralgraybg}{HTML}{F7F8FA}

\newtcolorbox{headerbox}[3][]{%
    enhanced,
    breakable,
    colback=#3,
    colframe=#2!70!black,
    boxrule=0.6pt,
    arc=3pt,
    left=8pt,
    right=8pt,
    top=11pt,
    bottom=7pt,
    before skip=4pt,
    after skip=2pt,
    fontupper=\small,
    title={#1},
    fonttitle=\bfseries\small,
    coltitle=black,
    attach boxed title to top left={xshift=8pt,yshift=-2.2mm},
    boxed title style={
        colback=#2!18!white,
        colframe=#2!18!white,
        boxrule=0pt,
        arc=2pt,
        left=5pt,
        right=5pt,
        top=2pt,
        bottom=2pt
    }
}


\newtcolorbox{boxblue}{enhanced,colback=blue!5!white,colframe=blue!75!black,breakable=true}

\usepackage{listings}

\lstdefinelanguage{json}{
    basicstyle=\scriptsize\ttfamily,
    numbers=none,
    numberstyle=\tiny,
    stepnumber=1,
    numbersep=5pt,
    showstringspaces=false,
    breaklines=true,
    frame=none,
    backgroundcolor=\color{white},
    literate=
     *{:}{{{\color{black}{:}}}}{1}
      {,}{{{\color{black}{,}}}}{1}
      {\{}{{{\color{black}{\{}}}}{1}
      {\}}{{{\color{black}{\}}}}}{1}
      {[}{{{\color{black}{[}}}}{1}
      {]}{{{\color{black}{]}}}}{1},
}

\definecolor{YaleBlue}{RGB}{0, 53, 107}
\definecolor{UChiRed}{RGB}{128, 0, 0}

\newcommand{\UChi}{\hspace{.1em}^{\textcolor{UChiRed}{\boldsymbol{C}}}}
\newcommand{\Yale}{\hspace{.1em}^{\textcolor{YaleBlue}{\boldsymbol{Y}}}}

\newcommand{\huggingface}{\raisebox{-1.5pt}{\includegraphics[height=1.05em]{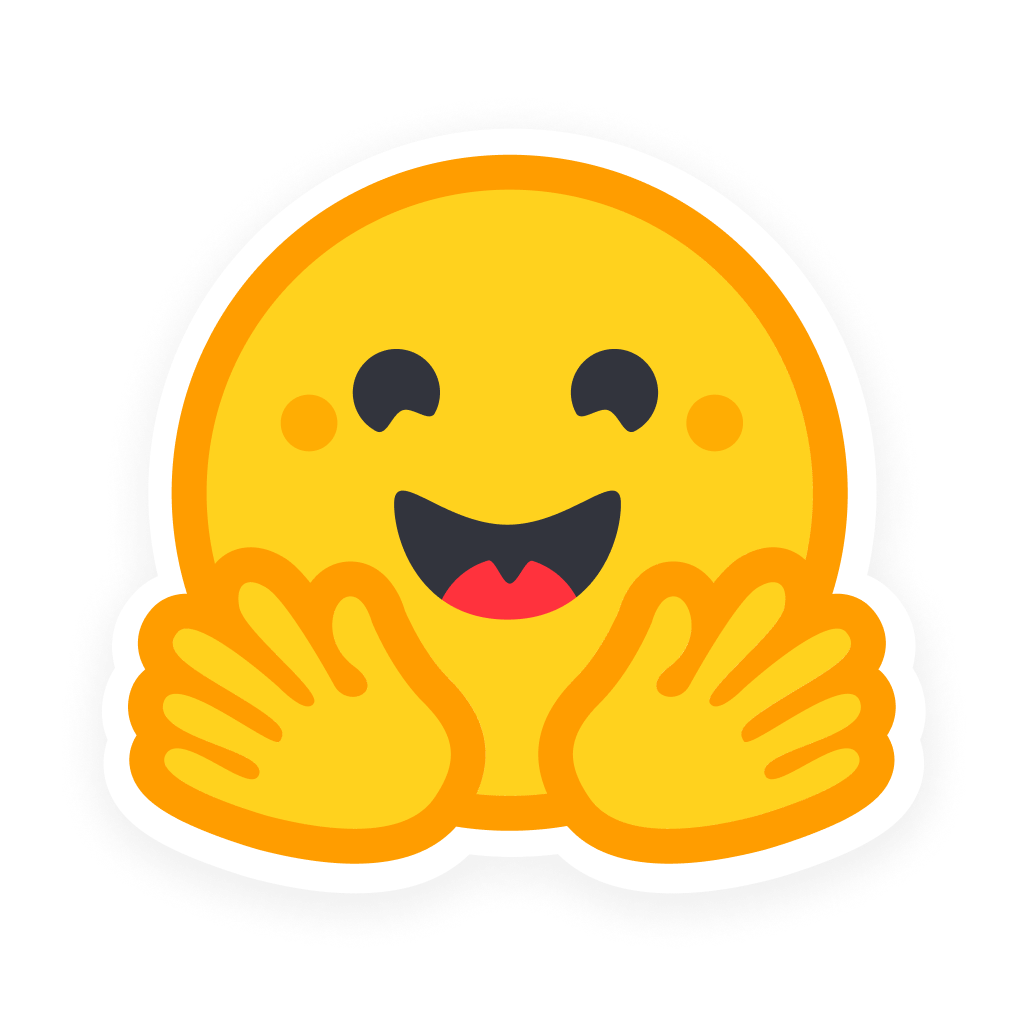}}\xspace}
\newcommand{\github}{\raisebox{-1.5pt}{\includegraphics[height=1.05em]{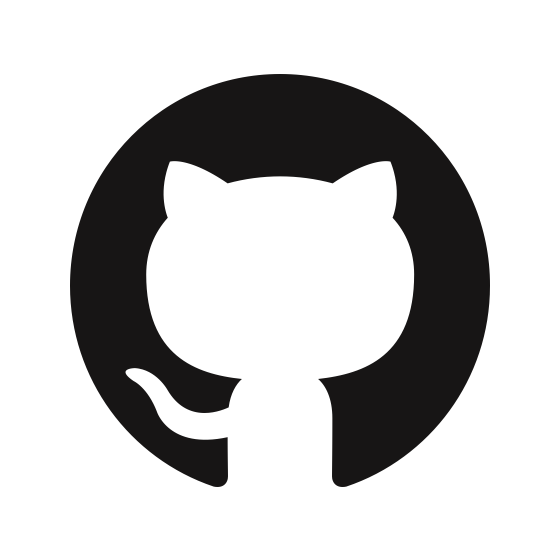}}\xspace}

\title{\fontsize{15}{17}\selectfont Measuring the Gap Between Human and LLM Research Ideas}

\author{
\textbf{Ziyu Chen}$\UChi$ \qquad
\textbf{Yilun Zhao}$\Yale$ \qquad
\textbf{Arman Cohan}$\Yale$\\ [4pt]
$\Yale$Yale University \qquad $\UChi$University of Chicago\\ [8pt]
\href{https://github.com/ziyuuc/TasteGap}{\github \texttt{ziyuuc/TasteGap}} \qquad
\href{https://huggingface.co/datasets/idealand/IdeaSeed}{\huggingface \texttt{IdeaLand/IdeaSeed}}\\[4pt]
}

\begin{document}

\maketitle
\thispagestyle{fancy}
\fancyhead{}
\lhead{%
    \includegraphics[height=1.1cm]{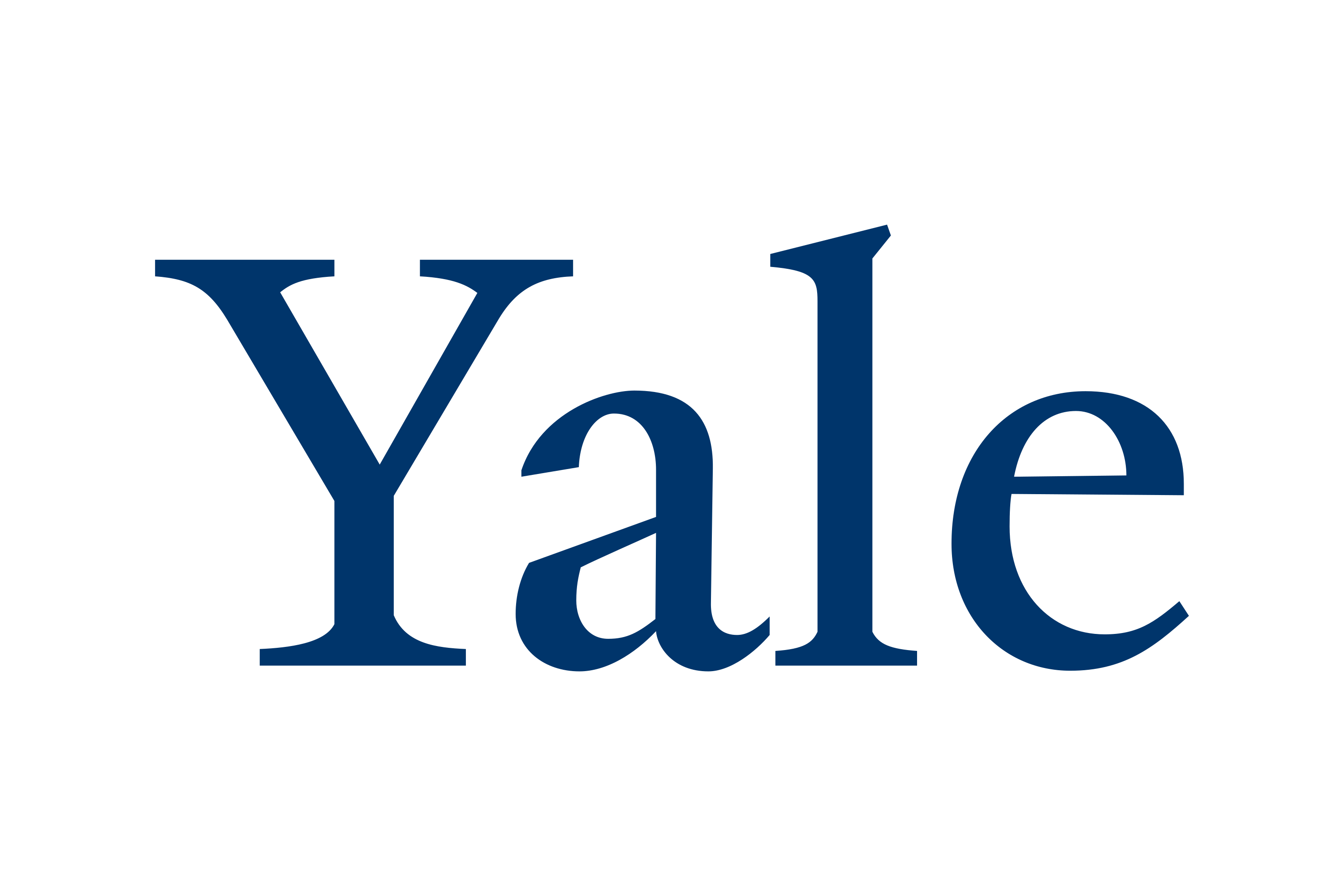}\hspace{0.2cm}%
}
\rhead{%
    \includegraphics[height=1.05cm]{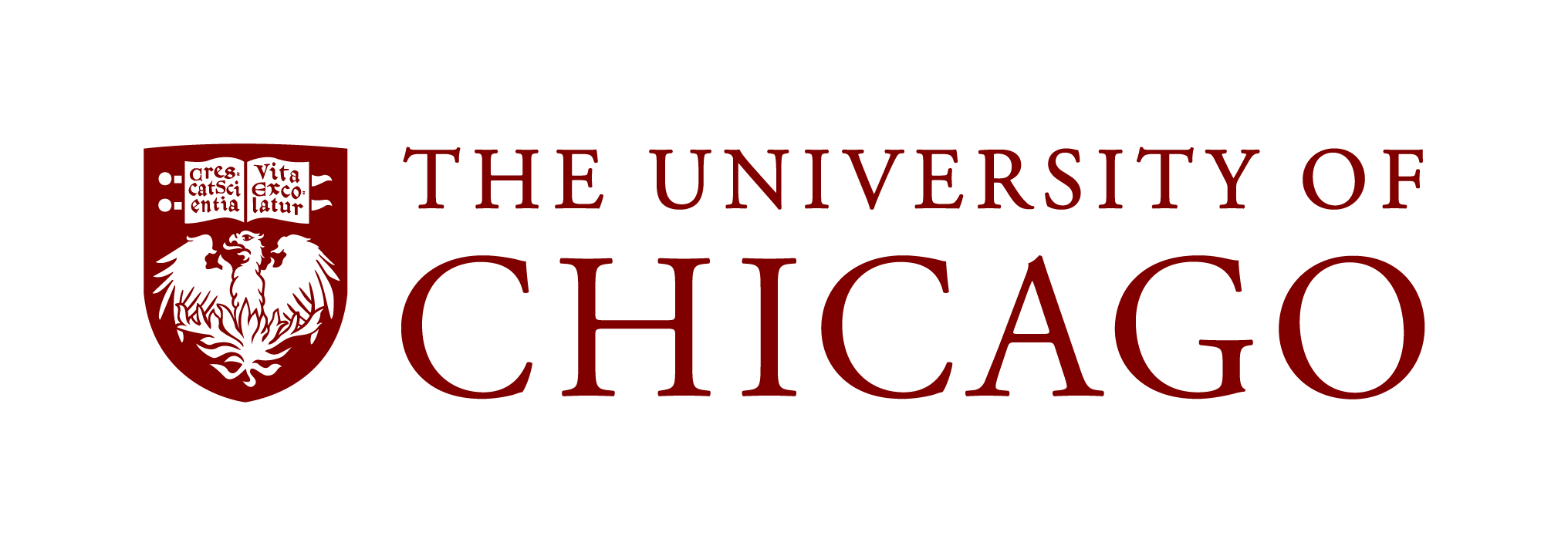}\hspace{0.2cm}%
}

\vspace{0.5em}

\fancyfoot[C]{\thepage}
\renewcommand{\headrulewidth}{0pt}
\setlength{\headheight}{12pt}
\addtolength{\topmargin}{0pt}
\setlength{\headsep}{3mm}

\vspace{-0.5em}

\pagestyle{plain}
\setlength{\parskip}{0.5ex plus 0.2ex minus 0.1ex}

\begin{abstract}
LLMs are increasingly used to brainstorm research ideas, but existing evaluations mostly judge individual ideas by novelty, feasibility, or expert preference. 
We instead ask: how far are current LLM-generated ideas from human researchers? To characterize this gap, we build a large-scale evaluation framework for ideation from high-quality human research papers. For each paper, we reverse-engineer a small set of closely related prior works that likely inspired its core idea. LLMs are then prompted to generate a new idea from the set of paper titles and summaries.
We introduce a two-axis research-taste taxonomy to profile each idea by its opportunity pattern and research paradigm, and use it to quantify the divergence between human and LLM ideas. 
Across idea sets generated by different LLMs, we observe a consistent distributional gap: LLM ideas are disproportionately concentrated around bridge-like opportunities and synthesis methods, whereas the human paper reference distribution spreads more broadly across ways of framing gaps and constructing contributions.
This result suggests that strong LLMs can produce a range of reasonable ideas, but that range remains narrower than, and systematically shifted relative to, human research taste.

\end{abstract}

\begin{figure}[h]
\centering
\includegraphics[width=0.7\textwidth]{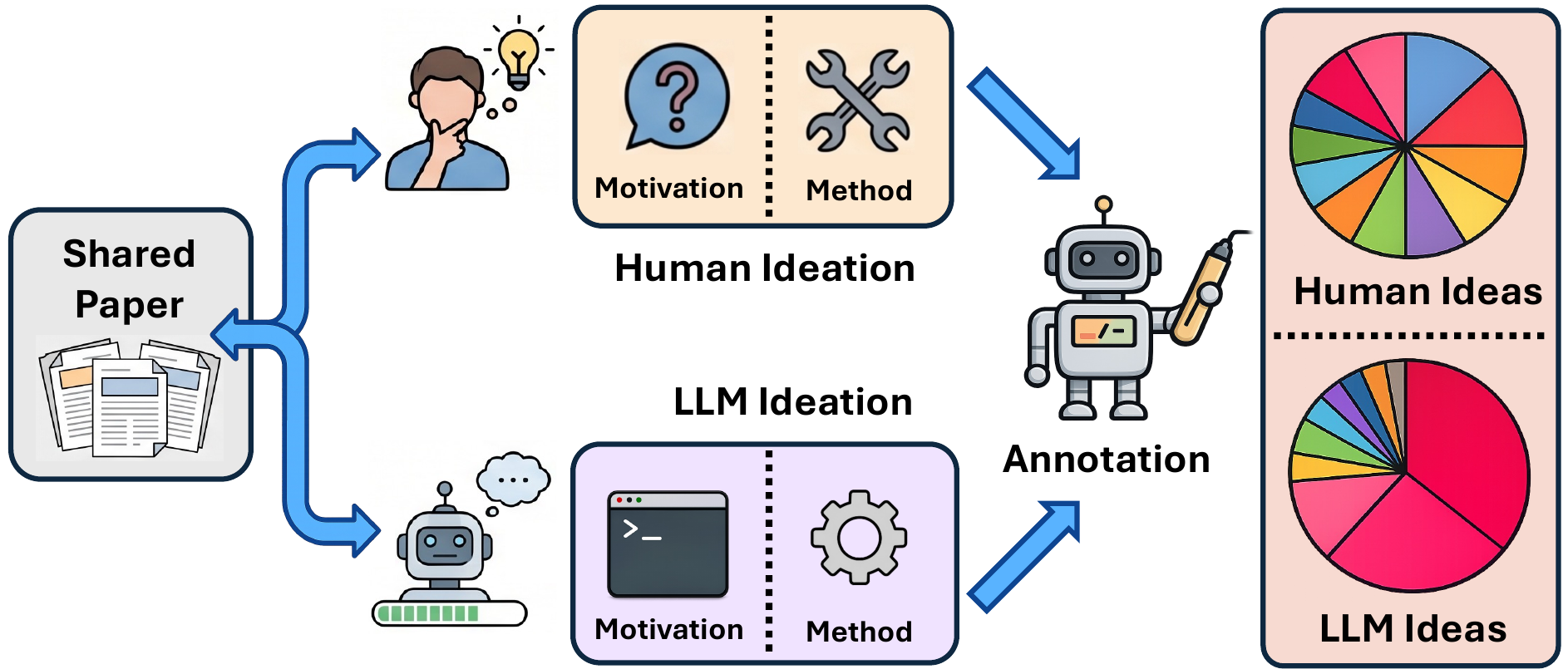}
\caption{
Overview of our research-taste gap analysis. From a shared literature context, humans contribute the paper idea while LLMs generate new ideas from the same prior works. Each idea is decomposed into a \textit{motivation} and a \textit{method}, then annotated with a research-taste taxonomy. Comparing the resulting distributions reveals that LLM ideas are substantially narrower than human ideas, with strong biases toward bridge-like motivations and explicit synthesis methods.
}
\label{fig:intro}
\end{figure}

\section{Introduction}

Research ideation is one of the most ambitious proposed uses of LLMs. Some controlled human studies have shown that LLM-generated ideas can match or approach those from human experts in terms of judged novelty and feasibility, demonstrating the potential of LLMs as ideation tools~\citep{si2024llmideas,baek2024researchagent,si2025ideation}. More recently, AI-scientist systems have already begun adapting LLMs to generate research ideas, execute experiments, and write paper drafts, bringing this capability into scientific process~\citep{boiko2023autonomous, lu2024aiscientist,zhang2024comprehensive,zhao-etal-2025-abgen,vasu2025hyper,zhao2026sciarena, wu-etal-2026-rbtact}. Despite this progress, a basic empirical question remains unanswered: \textit{What kinds of research ideas do LLMs tend to produce, and how far are current LLM-generated ideas from human researchers?}

Most existing evaluations of LLM ideation judge ideas individually, using criteria such as novelty, feasibility, impact, or preference~\citep{si2024llmideas, baek2024researchagent,garikaparthi-etal-2025-iris,tong2026ailearnscientifictaste}. In this work, we take a complementary distributional view. We use \textit{research taste} to refer to the kinds of problems, gaps, and contributions that a source tends to produce across many comparable literature-grounded ideation contexts. Under this view, research taste concerns not only whether an individual idea is reasonable, but also what kinds of gap framings and contribution strategies repeatedly appear when the same source is asked to generate ideas under comparable constraints.

This distributional view matters because a single idea may appear novel, feasible, and coherent, while the broader set of ideas from the same source may still reflect a narrow range of research taste. Research communities generate many kinds of contributions: some papers discover a failure mode, some relax an assumption, some build a measurement instrument, some introduce a formal explanation, and others construct a system or artifact. An LLM that generates reasonable ideas one at a time may therefore still be behaviorally narrow if its outputs repeatedly identify the same kinds of gaps, use the same methodological paradigms, or rely on the same contribution templates~\citep{meincke2024prompting,smith2025comprehensive,sourati2026homogenizing}. Such concentration would affect how LLMs are used for brainstorming, literature exploration, and automated research agents, even when many individual outputs appear coherent.

We study this question through a constrained literature-grounded ideation task.
Each example consists of a small set of closely related prior papers, represented by their titles and abstracts. The target output is a new research idea, separated into a \textit{motivation} and a \textit{method}. 
This differs from open-ended ideation such as \textit{write an idea about topic X}. Grounding each generation in a small related-work context makes the task comparable across human and LLM outputs. The human idea is the idea realized in the real paper, while the LLM idea is generated from the reconstructed local context that seemingly reasonably preceded that paper. This setting also reduces the chance that differences are driven only by broad topic choice or by an LLM's preferred generic paper format.

To build the evaluation framework, we collect real papers in machine learning and natural science domains. For each paper, we use a strong LLM-assisted extraction pipeline to identify the paper's core human idea and to reverse-engineer several closely related prior works from which that idea can be understood. We retrieve these prior works and prompt evaluated LLMs to produce a new motivation and method from the same prior-work context.
This yields paired human and LLM idea corpora over the same inputs.

We compare these corpora at the distributional level through a two-dimensional view of research taste. One dimension characterizes how a proposal frames the underlying research opportunity, ranging from identifying missing explanations or overlooked failures to exposing structural disconnects or limitations in existing understanding. The other captures the style of intellectual contribution through which the opportunity is developed into a research idea, including analytical, constructive, integrative, and exploratory forms of proposed methods. 
We introduce a taxonomy along these two axes, constructed by human experts through a review of research guidance from NSF, NIH, AHRQ and DARPA, and then iteratively refined using a held-out set of papers to ensure applicability across both machine learning and natural science domains.
We apply the taxonomy at scale using an LLM annotator validated against independent human judgments.

We find that LLM-generated ideas occupy a substantially narrower region of the research-taste taxonomy than human ideas. This narrowing is most visible in an ideation pattern centered on connection, where model ideas more often frame the motivation as a need to link previously literatures, methods, or evidence streams, and more often develop the method by integrating, reconciling, or unifying existing approaches. In our evaluation, only 12.1\% of human ideas motivated by the pattern of connection, and only 5.1\% use synthesis or unification as the central method paradigm. By contrast, across the nine main evaluated LLMs, the corresponding rates range from 47.1\% to 64.2\% and from 22.5\% to 38.7\%, respectively. Human ideas also exhibit consistently higher normalized entropy on both taxonomy axes. This pattern remains stable across model families and scientific domains, indicating that current LLM ideation is disproportionately concentrated around integrative and synthesis-oriented types, while human research ideas span a substantially broader range of opportunity patterns and methodological paradigms.

\section{Related Work}

\paragraph{LLMs for Research Ideation.}
Recent work has explored LLMs for scientific ideation, including generating, refining, and evaluating research hypotheses and directions. Early studies show that directly prompted LLMs can produce ideas perceived as highly novel, though often less feasible or well-grounded than human proposals~\citep{si2024llmideas}. Building on this, subsequent work has explored iterative refinement, retrieval-augmented generation, and search-based ideation pipelines~\citep{wang2024scimon,baek2024researchagent,sanyal2025spark}. Other approaches ground generation in external scientific signals such as retrieved literature, knowledge graphs, or emerging research trends~\citep{ghafarollahi2024sciagents,hu2024nova,pu2025ideasynth}. Multi-agent collaboration has also become a common paradigm for improving idea diversity and critique through simulated scientific discussion~\citep{gu2024combinatorial,su2024virsci}. Benchmarks are proposed to evaluate dimensions including novelty, feasibility, and impact of generated ideas~\citep{liveideabench2024,guo2025ideabench,researchbench2025}. Our work differs from this line of research by studying whether the overall \textit{distribution} of LLM-generated ideas resembles the distribution of ideas realized in human-written scientific papers.

\paragraph{Gaps between Human and LLM-Generated Content.}
Even when LLM outputs appear fluent and useful, research shows that they can differ systematically from human outputs. Early detection work found that neural generations contain statistical artifacts in token ranks, sampling behavior, and likelihood geometry that distinguish them from human text~\citep{gehrmann-etal-2019-gltr,ippolito-etal-2020-automatic,mitchell2023detectgpt}; broader comparison corpora and distributional metrics similarly show measurable gaps between ChatGPT or neural text and human expert writing~\citep{pillutla2021mauve,guo2023close}. These gaps become more consequential when LLM outputs are used as substitutes for human populations or human judgments. In social simulation, LLMs can reproduce some aggregate patterns while still exhibiting distortions or demographic misalignment relative to real human responses~\citep{aher2023simulate,santurkar2023whose}. In evaluation and review settings, LLM judges vary substantially across tasks and require validation against human annotations, while LLM-generated paper reviews over-focus on technical validity and under-attend to novelty compared with expert reviewers~\citep{bavaresco-etal-2025-llms,shin-etal-2025-mind}. Our work follows this perspective of comparison between human and LLMs, shifting the target to the distribution of research ideas.

\section{Evaluation Framework for Ideation}
\label{sec:method}

We propose an idea-level evaluation framework to systematically analyze what kinds of research opportunities and contribution strategies LLMs emphasize or overlook during scientific ideation. We first define a literature-grounded ideation task and construct paired human and LLM idea corpora accordingly. We represent each idea through its motivation and method, annotate these ideas with a research-taste taxonomy, and analyze.

\subsection{Literature-Grounded Ideation Task}

Each instance contains a set of related prior works
$X_i=\{(t_{i1},a_{i1}),\ldots,(t_{ik},a_{ik})\}$, with $t$ denoting the title and $a$ the abstract. The target output is a research idea \(y_i=(m_i,s_i)\), where \(m_i\) refers to the motivation and \(s_i\) stands for the proposed method. Guided by the provided literature context, the prompt directs models to identify research gaps across papers and generate a coherent research idea.
This task is constrained. In an open-ended ideation setting, differences between human and model outputs can be confounded by topic selection, prior knowledge, and generic paper-writing templates, making the comparison difficult to interpret~\citep{si2024llmideas,liveideabench2024}. By anchoring both human and model ideas to the same set of related prior works, we focus the analysis on how each source identifies research gaps and constructs contributions from a shared local context.

\subsection{Idea Corpus}
\label{sec:ideacorpus}

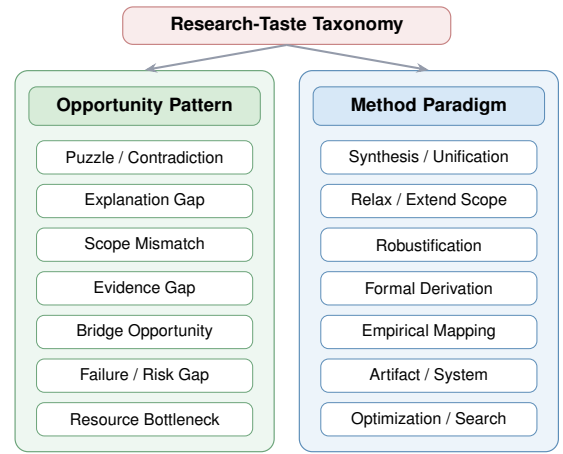
\begin{wrapfigure}{r}{0.45\textwidth}
    \centering
    \vspace{-15pt}
    \begin{adjustbox}{width=\linewidth}
    \begin{tikzpicture}[
        font=\sffamily,
        root/.style={
            rounded corners=4pt,
            fill=TaxRedBg,
            draw=TaxRedLine,
            line width=0.45pt,
            text width=4.15cm,
            inner sep=4pt,
            align=center,
            font=\scriptsize\bfseries\sffamily
        },
        panel/.style={
            rounded corners=5pt,
            line width=0.45pt,
            minimum width=3.50cm,
            minimum height=5.15cm,
            inner sep=0pt
        },
        paneltitle/.style={
            rounded corners=3pt,
            line width=0.4pt,
            text width=2.95cm,
            minimum height=0.52cm,
            inner sep=2.5pt,
            align=center,
            font=\scriptsize\bfseries\sffamily
        },
        taxnode/.style={
            rounded corners=3pt,
            fill=white,
            line width=0.35pt,
            minimum width=2.92cm,
            minimum height=0.45cm,
            inner xsep=1.8pt,
            inner ysep=1.6pt,
            align=center,
            font=\fontsize{6.25}{6.8}\selectfont\sffamily
        },
        link/.style={
            -{Stealth[length=1.7mm,width=1.3mm]},
            line width=0.8pt,
            draw=TaxSlate!70
        }
    ]
        \node[root] (root) at (0,0) {Research-Taste Taxonomy};

        \node[panel, fill=TaxGreenBg, draw=TaxGreenLine, anchor=north] (oppbox) at (-1.92,-0.6) {};
        \node[paneltitle, fill=TaxGreenHead, draw=TaxGreenLine, anchor=north] (opphead) at (-1.92,-0.82)
        {Opportunity Pattern};
        \node[taxnode, draw=TaxGreenLine] at (-1.92,-1.78) {Puzzle / Contradiction};
        \node[taxnode, draw=TaxGreenLine] at (-1.92,-2.37) {Explanation Gap};
        \node[taxnode, draw=TaxGreenLine] at (-1.92,-2.96) {Scope Mismatch};
        \node[taxnode, draw=TaxGreenLine] at (-1.92,-3.55) {Evidence Gap};
        \node[taxnode, draw=TaxGreenLine] at (-1.92,-4.14) {Bridge Opportunity};
        \node[taxnode, draw=TaxGreenLine] at (-1.92,-4.73) {Failure / Risk Gap};
        \node[taxnode, draw=TaxGreenLine] at (-1.92,-5.32) {Resource Bottleneck};

        \node[panel, fill=TaxBlueBg, draw=TaxBlueLine, anchor=north] (parabox) at (1.92,-0.6) {};
        \node[paneltitle, fill=TaxBlueHead, draw=TaxBlueLine, anchor=north] (parahead) at (1.92,-0.82)
        {Method Paradigm};
        \node[taxnode, draw=TaxBlueLine] at (1.92,-1.78) {Synthesis / Unification};
        \node[taxnode, draw=TaxBlueLine] at (1.92,-2.37) {Relax / Extend Scope};
        \node[taxnode, draw=TaxBlueLine] at (1.92,-2.96) {Robustification};
        \node[taxnode, draw=TaxBlueLine] at (1.92,-3.55) {Formal Derivation};
        \node[taxnode, draw=TaxBlueLine] at (1.92,-4.14) {Empirical Mapping};
        \node[taxnode, draw=TaxBlueLine] at (1.92,-4.73) {Artifact / System};
        \node[taxnode, draw=TaxBlueLine] at (1.92,-5.32) {Optimization / Search};

        \draw[link] (root.south) -- (oppbox.north);
        \draw[link] (root.south) -- (parabox.north);
    \end{tikzpicture}
    \end{adjustbox}
    \caption{The two-axis research-taste taxonomy. The opportunity pattern axis labels \textit{why} a new study is needed and captures a research gap, while the method paradigm axis labels \textit{how} the proposed work turns that gap into a contribution.}
    \label{fig:taxonomy}
    \vspace{-9pt}
\end{wrapfigure}

\paragraph{Human Idea.}
We build our evaluation data from published research papers drawn from two sources: machine learning conference proceedings from ICLR, ICML and NeurIPS published between 2023 and 2026 and Nature Communications released from 2023 to 2025, covering 71 major scientific disciplines such as physics, chemistry, and biology. For each paper, the work itself represents the human endpoint, which is the research idea originally devised by authors for publication. We extract this idea into a structured representation using an LLM-assisted pipeline. The extraction prompt asks for the innovation, departure from prior work, and key insight, then rewrites the result into a proposal-style motivation and method.
We then reverse-engineer 4 to 8 highly relevant prior studies based on the extracted idea and the paper's related-work section. For each related work, we retrieve the title and abstract. The final input contains only the prior-work titles and abstracts as main idea summaries. The mixed corpus contains 11,683 valid human ideas in total. Full prompts and additional data details are provided in~\autoref{app:dataset}.

\paragraph{LLM Idea.}

For each input, we prompt LLMs to generate a new research idea in the same structured format with two components. The motivation synthesizes research gaps across the provided papers, and the method outlines a concrete high-level approach. We evaluate a set of mainstream LLM families including \texttt{Claude}, \texttt{Gemini}, \texttt{GPT}, \texttt{DeepSeek} and \texttt{Qwen}.

\subsection{Research-Taste Taxonomy}

We label each idea with a two-axis research-taste taxonomy matching an idea's structure of \textit{motivation} and \textit{method}. The \textit{opportunity pattern} axis corresponds to the motivation: it asks what kind of gap makes the proposal worth pursuing, such as a contradiction, missing explanation, scope mismatch, evidence gap, disconnected literature, failure risk, or resource bottleneck. The \textit{method paradigm} axis corresponds to the method: it asks what high-level contribution strategy turns that gap into a paper, such as synthesis, scope extension, robustification, formal derivation, empirical mapping, artifact construction, or optimization.

We designed this taxonomy to support interpretable distributional comparison. We first reviewed guidance from NSF, NIH, AHRQ, and DARPA on research proposals and problem formulation. From these sources, we extracted recurring elements organized around two broad dimensions: the opportunity that motivates a study and the method by which the study addresses it. This initial taxonomy contained 11 opportunity elements and 9 method elements. We refined these candidate elements using a held-out set of 150 papers. For each paper, we allowed up to two closest labels on each axis and included an \textit{other} option for cases not covered by the initial taxonomy. We then examined coverage and label distributions, merged near-duplicate labels, split labels that conflated motivation and method, removed labels tied to a specific domain or technical substrate, and renamed categories using terms applicable across ML and natural-science papers. The final taxonomy contains seven opportunity patterns and seven method paradigms as shown in \autoref{fig:taxonomy}. Our final taxonomy meets three requirements: categories correspond to recurring gap framings and contribution strategies, 
generalizable across different research domains, 
and were checked during human validation to avoid systematic category collapse.
This ensures frequency discrepancies stem from distinct research tendencies, not field differences or wording variations. We provide details in \autoref{app:taxonomy}.

\subsection{Automated Annotation}
\label{sec:annotation}
We use an LLM as an automated annotator to label every human and LLM idea~\citep{zheng2023judging,liu2023g,kim2024prometheus}. The annotator receives the taxonomy, the prior-work titles for context, and the proposal motivation and method. It returns a primary and secondary label for each axis, confidence scores, and three diagnostic scores: surface stitching, bottleneck specificity, and boilerplate. We use primary labels for distributional comparisons and diagnostic scores for mechanism analyses.

Before applying the annotator at scale, we validated it on the same held-out set of 150 papers used for taxonomy calibration. Two authors examined ideas from this set and audited three outputs from each annotation pass including the opportunity-pattern label, the method-paradigm label, and the diagnostic-score profile. As a label-level reliability measure, we compute Cohen's \(\kappa\)~\citep{cohen1960coefficient} between the \texttt{GPT-5.4-mini}\footnote{We chose \texttt{GPT-5.4-mini} as it provided an appropriate balance between quality and cost.} annotation and each author's judgment, then average over the two human-LLM pairs. The resulting \(\kappa\) are 0.84, 0.81, and 0.93, respectively. We also inspect disagreements and confusion matrices to ensure that errors concentrate in semantically adjacent labels rather than reflecting systematic category collapse. Given this high agreement, we use \texttt{GPT-5.4-mini} as the automatic annotator for evaluation.

\section{Experiments}
\label{sec:experiments}

Our experiments ask whether LLMs generate the same \textit{kinds} of research ideas as human researchers when both are conditioned on the same local literature context. We first compare human and model label distributions on the two taxonomy axes. We then use the diagnostic scores to examine whether the same gap appears as lower specificity or more template-like proposals. Finally, we study how reasoning changes the distribution and use mechanism analyses to probe why the gap appears.

\subsection{Setup}

\paragraph{Data.}
We evaluate on the data introduced in~\autoref{sec:ideacorpus}. The full evaluation set contains 11,683 ground-truth human ideas from diverse fields mixed together, one per source paper, matched to generations from each evaluated model. We present overall results, while detailed analyses and results for each individual domain are provided in~\autoref{app:domain_results}.

\paragraph{Models.}
The main comparison includes nine model settings: \texttt{Claude-Sonnet-4.6}, \texttt{Gemini-3.1-Pro}, \texttt{GPT-OSS-20B}, \texttt{GPT-OSS-120B}, \texttt{GPT-5.4-mini}, \texttt{Qwen3-8B}, \texttt{Qwen3-32B}, \texttt{DeepSeek-V4-Flash}, and \texttt{DeepSeek-V4-Pro}~\citep{claudes46,gemini31,openai2025gptoss120bgptoss20bmodel,gpt54,yang2025qwen3,deepseekai2026deepseekv4}. For the reasoning ablation, we additionally evaluate \texttt{Qwen3-8B} and \texttt{DeepSeek-V4-Flash} with thinking mode. Each model is prompted with the same input and asked to produce a structured idea.

\paragraph{Metrics.}
For each source and taxonomy axis, we estimate the empirical distribution of primary labels. 
We use three distributional measures to characterize how model-generated ideas differ from human ideas. Total Variation Distance (TVD) measures the amount of label mass that would need to be reallocated for a model distribution to match the corresponding human distribution. Jensen-Shannon Divergence (JSD)~\citep{lin1991divergence} provides a bounded and symmetric measure of divergence between the model and human distributions. And normalized entropy, which measures how concentrated a source is over taxonomy labels. We report TVD as,
\[
\mathrm{TVD}(\hat{P},\hat{Q})=\frac{1}{2}\sum_{c\in A}|\hat{P}(c)-\hat{Q}(c)|,
\]
where \(\hat{P}\) and \(\hat{Q}\) denote empirical label distributions over the label set \(A\), and \(c\) indexes individual labels. We report JSD using base-2 logarithms,
\[
\mathrm{JSD}(\hat{P},\hat{Q})
=\frac{1}{2}\mathrm{KL}(\hat{P}\Vert M)
+\frac{1}{2}\mathrm{KL}(\hat{Q}\Vert M),
\]
where \(M=(\hat{P}+\hat{Q})/2\). Normalized entropy over the labels
\[
H_{\mathrm{norm}}(\hat{P}) =
-\frac{1}{\log_2 |A|}\sum_{c\in A}\hat{P}(c)\log_2 \hat{P}(c).
\] Lower normalized entropy indicates greater concentration over a smaller set of research moves~\citep{sajjadi2018assessing,pillutla2021mauve}.

\subsection{Main Distributional Gap}

\begin{table}[t]
    \centering
    \setlength{\tabcolsep}{12pt}
    \small
    \caption{Distributional distance between human and LLM ideas on our evaluation set. \textbf{Opportunity Pattern} captures why the idea is worth pursuing, while \textbf{Method Paradigm} captures how the proposal turns that gap into a contribution. Ent. is normalized entropy. All model distributions remain far from the human distribution, especially on the opportunity axis. \raisebox{0.1ex}{\textcolor{blue!35}{\rule{0.65em}{0.65em}}} marks the best non-human score, and \raisebox{0.1ex}{\textcolor{red!35}{\rule{0.65em}{0.65em}}} marks the clearest model-side degradation.}
    \label{tab:main_gap}
    \begin{tabular}{lcccccc}
        \toprule
        & \multicolumn{3}{c}{\textbf{Opportunity Pattern}} & \multicolumn{3}{c}{\textbf{Method Paradigm}} \\
        \cmidrule(lr){2-4}\cmidrule(lr){5-7}
        Source & TVD \(\downarrow\) & JSD \(\downarrow\) & Ent. \(\uparrow\) & TVD \(\downarrow\) & JSD \(\downarrow\) & Ent. \(\uparrow\) \\
        \midrule
        \rowcolor{black!3}
        Human & --- & --- & 0.926 & --- & --- & 0.920 \\
        \midrule
        \texttt{Claude-Sonnet-4.6} & 0.351 & 0.130 & 0.737 & \cellcolor{blue!8}\textbf{0.211} & \cellcolor{blue!8}\textbf{0.070} & \cellcolor{blue!8}\textbf{0.879} \\
        \texttt{Gemini-3.1-Pro} & \cellcolor{blue!8}\textbf{0.348} & \cellcolor{blue!8}\textbf{0.128} & \cellcolor{blue!8}\textbf{0.758} & 0.227 & 0.092 & 0.874 \\
        \texttt{GPT-OSS-20B} & 0.456 & 0.218 & 0.598 & 0.378 & 0.158 & \cellcolor{red!8}0.723 \\
        \texttt{GPT-OSS-120B} & \cellcolor{red!8}0.521 & \cellcolor{red!8}0.259 & \cellcolor{red!8}0.550 & \cellcolor{red!8}0.391 & 0.170 & 0.735 \\
        \texttt{GPT-5.4-mini} & 0.512 & 0.243 & 0.568 & 0.339 & 0.119 & 0.814 \\
        \texttt{Qwen3-8B} & 0.382 & 0.179 & 0.658 & 0.368 & \cellcolor{red!8}0.190 & 0.734 \\
        \texttt{Qwen3-32B} & 0.417 & 0.191 & 0.640 & 0.364 & 0.183 & 0.745 \\
        \texttt{DeepSeek-V4-Flash} & 0.400 & 0.167 & 0.683 & 0.246 & 0.086 & 0.845 \\
        \texttt{DeepSeek-V4-Pro} & 0.436 & 0.208 & 0.642 & 0.258 & 0.108 & 0.828 \\
        \bottomrule
    \end{tabular}
\end{table}

\begin{figure}[!t]
    \centering
    \includegraphics[width=0.90\linewidth]{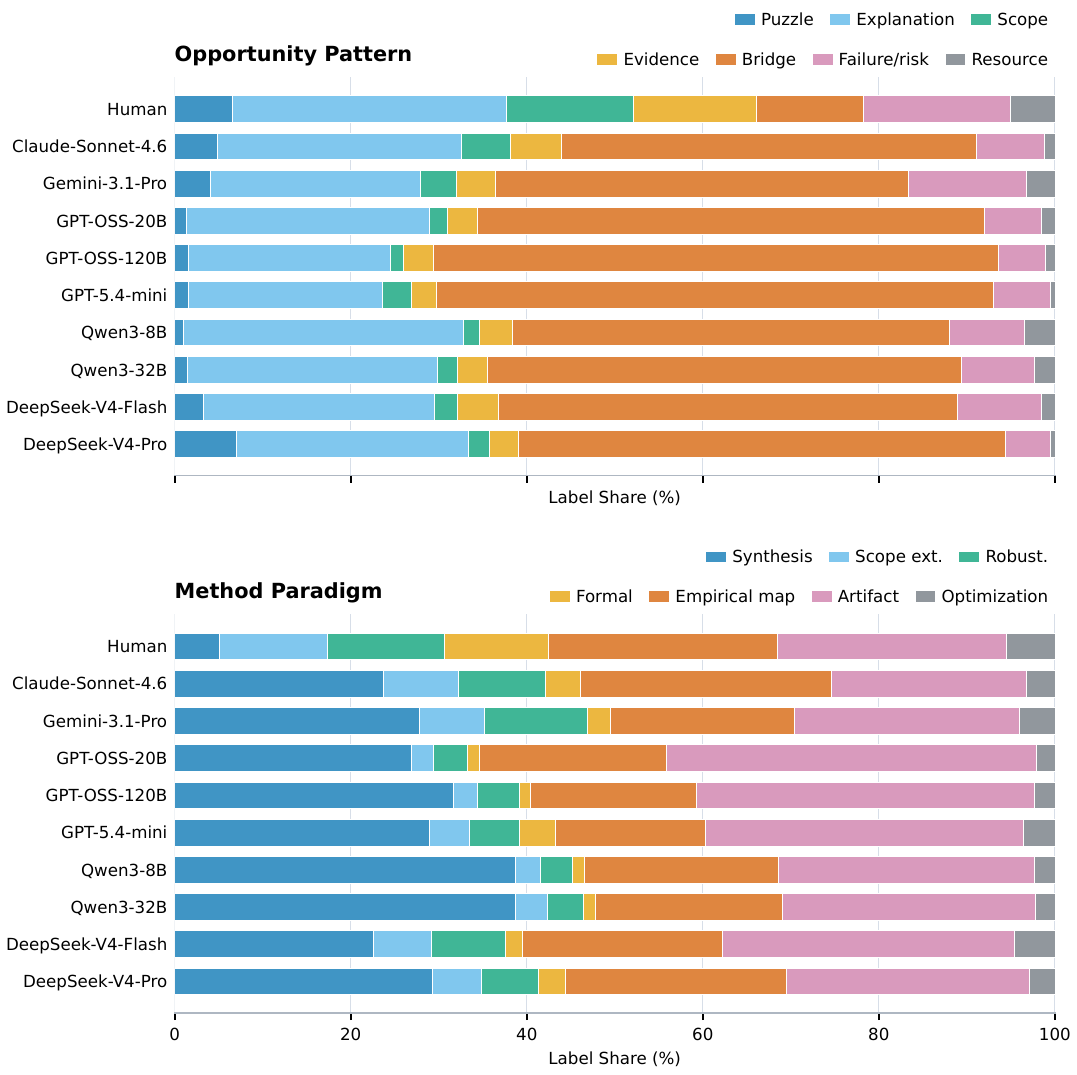}
    \caption{Full label distributions over opportunity patterns and method paradigms. On the opportunity axis, LLMs strongly amplify fragmentation and bridge opportunities (\raisebox{0.1ex}{\textcolor[RGB]{224,134,64}{\rule{0.65em}{0.65em}}}), where the research gap is framed as disconnected literatures, methods, or evidence streams that should be connected. On the method axis, the largest shift is toward explicit synthesis or unification (\raisebox{0.1ex}{\textcolor[RGB]{64,149,197}{\rule{0.65em}{0.65em}}}), where the contribution is constructed by integrating or reconciling separate ideas. Thinking-mode outputs are shown for \texttt{Qwen3-8B} and \texttt{DeepSeek-V4-Flash}.}
    \label{fig:main_distribution}
\end{figure}

\autoref{fig:main_distribution} and \autoref{tab:main_gap} show a consistent distributional gap across tested model families. Human ideas have high normalized entropy on both taxonomy axes, above 0.92. Model distributions are generally more concentrated than the human reference distribution, especially on the opportunity axis: opportunity entropy ranges from 0.550 to 0.758, while method-paradigm entropy ranges from 0.723 to 0.879. Even the closest model on the opportunity axis, \texttt{Gemini-3.1-Pro}, has TVD 0.348, meaning that over a third of the distributional mass would need to move to match human outputs. On the method axis, \texttt{Claude-Sonnet-4.6} is closest, but its TVD remains 0.211.

The largest gap is a shift toward bridge-and-synthesis ideas. Only 12.1\% of human opportunities are labeled as fragmentation or bridge opportunities, compared with 47.1 to 64.2\% for the main LLMs. The same pattern appears on the method axis: explicit synthesis or unification accounts for 5.1\% of human ideas but 22.5 to 38.7\% of LLM ideas. This does not mean that synthesis is never a valid research contribution. Rather, it shows that when models are given nearby papers, they frequently turn ideation into a generic move of connecting or combining prior work, while human papers distribute more mass to explanation, measurement, risk, scope, artifacts, and optimization-style contributions. 
The mechanism analyses in \autoref{sec:mechanisms} ask whether this gap reflects many diverse forms of synthesis or a narrower template for constructing proposals.
We also test alternative prompts, and this preference for organizing new ideas through bridging and synthesis does not appear sensitive to the specific wording of the generation prompt. Details in \autoref{app:prompt_ablation}.

\paragraph{Full-paper context ablation.}
To better approximate richer ideation settings, we also evaluate a full-paper context condition in which models can draw on the original related-work papers rather than only abstract summaries. In this setting, for each related work paper, LLMs first read the full text and produce their own summary; these model-generated summaries are then used in place of abstracts as the input context for idea generation. We run this ablation on a subset sampling 500 papers from each of the two domains and evaluate \texttt{Qwen3-8B} and \texttt{DeepSeek-V4-Flash}. \autoref{tab:full_context_main} shows that richer context does not move either model closer to the human reference distribution. For both models, TVD and JSD increase on both taxonomy axes under full-paper summaries, while entropy decreases or remains nearly unchanged. The full label distributions in \autoref{app:full_context} show the same qualitative pattern: even when models are given a richer representation of the related literature, their generated ideas continue to exhibit the same preference for organizing new proposals through bridging and synthesis.

\begin{table}[H]
    \centering
    \small
    \setlength{\tabcolsep}{6pt}
    \caption{Full-paper context ablation on a 1,000 paper subset. Full context replaces abstracts with model-generated full-paper summaries with detailed \textit{motivation}, \textit{method} and \textit{insight}. TVD / JSD are computed against the human reference distribution, and Ent. is normalized entropy. Full label counts are in \autoref{tab:full_context_ablation}.}
    \label{tab:full_context_main}
    \begin{adjustbox}{max width=\linewidth}
    \begin{tabular}{llllllll}
        \toprule
        & & \multicolumn{3}{c}{\textbf{Opportunity Pattern}} & \multicolumn{3}{c}{\textbf{Method Paradigm}} \\
        \cmidrule(lr){3-5}\cmidrule(lr){6-8}
        Model & Context & TVD \(\downarrow\) & JSD \(\downarrow\) & Ent. \(\uparrow\) & TVD \(\downarrow\) & JSD \(\downarrow\) & Ent. \(\uparrow\) \\
        \midrule
        \texttt{Qwen3-8B} & Abstract & 0.376 & 0.165 & 0.669 & 0.338 & 0.182 & 0.752 \\
        \texttt{Qwen3-8B} & Full & 0.430 {\scriptsize\textcolor{red!70!black}{(+.054)}} & 0.205 {\scriptsize\textcolor{red!70!black}{(+.040)}} & 0.623 {\scriptsize\textcolor{green!70!black}{(-.046)}} & 0.400 {\scriptsize\textcolor{red!70!black}{(+.062)}} & 0.229 {\scriptsize\textcolor{red!70!black}{(+.047)}} & 0.699 {\scriptsize\textcolor{green!70!black}{(-.053)}} \\
        \texttt{DeepSeek-V4-Flash} & Abstract & 0.368 & 0.152 & 0.706 & 0.213 & 0.079 & 0.867 \\
        \texttt{DeepSeek-V4-Flash} & Full & 0.400 {\scriptsize\textcolor{red!70!black}{(+.032)}} & 0.160 {\scriptsize\textcolor{red!70!black}{(+.008)}} & 0.701 {\scriptsize\textcolor{green!70!black}{(-.005)}} & 0.236 {\scriptsize\textcolor{red!70!black}{(+.023)}} & 0.093 {\scriptsize\textcolor{red!70!black}{(+.014)}} & 0.860 {\scriptsize\textcolor{green!70!black}{(-.007)}} \\
        \bottomrule
    \end{tabular}
    \end{adjustbox}
\end{table}

\subsection{Diagnostic Scores}

\begin{table}[t]
    \centering
    \caption{Diagnostic scores from the automatic annotator. The annotator assigns three ordinal 0 to 3 ratings from the proposal motivation and method: surface stitching measures whether the idea is a superficial combination, bottleneck specificity measures whether it identifies a precise mechanism or limiting factor, and boilerplate measures generality. \raisebox{0.1ex}{\textcolor{red!35}{\rule{0.65em}{0.65em}}} marks the clearest model-side degradation, and \raisebox{0.1ex}{\textcolor{blue!35}{\rule{0.65em}{0.65em}}} marks the best non-human diagnostic scores.}
    \label{tab:diagnostics}
    \small
    \begin{tabular}{lcccc}
        \toprule
        Source & \makecell{Surf. Score \(\downarrow\)} & \makecell{Surf. Flag (\%) \(\downarrow\)} & \makecell{Bottleneck \(\uparrow\)} & \makecell{Boilerplate \(\downarrow\)} \\
        \midrule
        \rowcolor{black!3}
        Human & \textbf{0.00} & \textbf{0.0} & \textbf{2.56} & \textbf{0.48} \\
        \midrule
        \texttt{Claude-Sonnet-4.6} & \cellcolor{blue!8}\textbf{0.02} & \cellcolor{blue!8}\textbf{0.1} & \cellcolor{blue!8}\textbf{2.60} & \cellcolor{blue!8}\textbf{0.37} \\
        \texttt{Gemini-3.1-Pro} & 0.09 & 0.4 & 2.34 & 0.79 \\
        \texttt{GPT-OSS-20B} & 0.09 & 1.1 & 2.07 & 0.97 \\
        \texttt{GPT-OSS-120B} & 0.07 & 0.3 & 2.16 & 0.87 \\
        \texttt{GPT-5.4-mini} & 0.02 & 0.1 & 2.21 & 0.75 \\
        \texttt{Qwen3-8B} & \cellcolor{red!8}0.58 & \cellcolor{red!8}20.6 & \cellcolor{red!8}1.76 & \cellcolor{red!8}1.25 \\
        \texttt{Qwen3-8B-Think} & 0.45 & 11.0 & 1.90 & 1.11 \\
        \texttt{Qwen3-32B} & 0.44 & 13.7 & 1.87 & 1.15 \\
        \texttt{DeepSeek-V4-Flash} & 0.10 & 1.2 & 2.12 & 0.92 \\
        \texttt{DeepSeek-V4-Flash-Think} & 0.10 & 0.7 & 2.12 & 0.89 \\
        \texttt{DeepSeek-V4-Pro} & 0.04 & 0.2 & 2.34 & 0.69 \\
        \bottomrule
    \end{tabular}
\end{table}

We next analyze the three diagnostic scores defined in \autoref{sec:annotation} and reported in \autoref{tab:diagnostics}: surface stitching, bottleneck specificity, and boilerplate. These annotator-assigned scores characterize the specificity and template-likeness of each proposal. Surface stitching measures whether the idea is a superficial combination of prior work, bottleneck specificity measures whether it identifies a precise mechanism or limiting factor, and boilerplate measures generic phrasing. In \autoref{tab:diagnostics}, most model outputs receive lower specificity and higher boilerplate scores than human ideas, with especially strong degradation for the \texttt{Qwen} models, which also have the highest surface-stitching scores and flags. \texttt{Claude-Sonnet-4.6} is an exception on these diagnostic dimensions: it has slightly higher bottleneck specificity and lower boilerplate than the human baseline, while still remaining distributionally shifted in \autoref{tab:main_gap}. 
Taken as supporting evidence, these annotator-based diagnostic scores suggest that the research-taste gap is not only a low-level quality issue;
even polished and specific model outputs can concentrate on a narrower set of opportunity and method patterns. Surface stitching is a stricter local judgment, while bridge and synthesis labels describe the broader research type. Together, however, they support the same interpretation: LLM ideation often converges on polished combinations of prior work, and for many models those combinations are also less specific. This interpretation is consistent with the archetype analysis in~\autoref{sec:mechanisms}, where model-heavy clusters have lower bottleneck specificity and higher boilerplate than human-heavy clusters built around replacing or decoupling a local mechanism~\citep{gehrmann-etal-2019-gltr,ippolito-etal-2020-automatic}.

\subsection{Does Extended Reasoning Help? }

Reasoning is considered a paradigm that can enhance the downstream capabilities of models~\citep{kojima2022large,wei2022chain,guo2025deepseek}. However, in our ideation tasks, thinking mode moves the output distribution farther from the human reference for both model settings we test as shown in \autoref{tab:think_ablation}.
For \texttt{Qwen3-8B}, enabling thinking increases bridge opportunities from 49.7\% to 71.1\% and explicit synthesis from 38.7\% to 52.2\%. Opportunity entropy drops from 0.658 to 0.481, and TVD from humans increases from 0.382 to 0.590. This adverse effect persists across both weaker and stronger models we tested. 
The same direction appears for \texttt{DeepSeek-V4-Flash}: bridge opportunities rise from 52.2\% to 59.1\%, synthesis rises from 22.5\% to 30.7\%, and both opportunity and method TVD increase. Thinking therefore appears to sharpen the model's preferred ideation template instead of broaden the distribution toward human taste, and further reduces the diversity of generated ideas.

\begin{table}[H]
    \centering
    \small
    \caption{Comparison between models with and without reasoning enabled, reporting changes relative to the corresponding base model. Enabling thinking consistently increases bridge and synthesis mass, increases TVD from the human distribution, and lowers entropy, indicating a sharper and less human-like idea distribution.}
    \label{tab:think_ablation}
    \resizebox{\textwidth}{!}{%
    \begin{tabular}{lllllllll}
        \toprule
        & \multicolumn{2}{c}{Template Mass} & \multicolumn{4}{c}{Distance to Human} & \multicolumn{2}{c}{Diagnostics} \\
        \cmidrule(lr){2-3}\cmidrule(lr){4-7}\cmidrule(lr){8-9}
        Setting & Bridge \(\downarrow\) & Synthesis \(\downarrow\) & Opp. TVD \(\downarrow\) & Opp. Ent. \(\uparrow\) & Meth. TVD \(\downarrow\) & Meth. Ent. \(\uparrow\) & Surface \(\downarrow\) & Boiler. \(\downarrow\) \\
        \midrule
        \texttt{Qwen3-8B} & 49.7 & 38.7 & 0.382 & 0.658 & 0.368 & 0.734 & 0.58 & 1.25 \\
        \hspace{0.8em}\texttt{w/ think} &
        71.1 {\scriptsize\textcolor{red!70!black}{(+21.4)}} &
        52.2 {\scriptsize\textcolor{red!70!black}{(+13.5)}} &
        0.590 {\scriptsize\textcolor{red!70!black}{(+.208)}} &
        0.481 {\scriptsize\textcolor{red!70!black}{(-.177)}} &
        0.472 {\scriptsize\textcolor{red!70!black}{(+.104)}} &
        0.649 {\scriptsize\textcolor{red!70!black}{(-.085)}} &
        0.45 {\scriptsize\textcolor{green!70!black}{(-.13)}} &
        1.11 {\scriptsize\textcolor{green!70!black}{(-.14)}} \\
        \midrule
        \texttt{DeepSeek-V4-Flash} & 52.2 & 22.5 & 0.400 & 0.683 & 0.246 & 0.845 & 0.10 & 0.92 \\
        \hspace{0.8em}\texttt{w/ think} &
        59.1 {\scriptsize\textcolor{red!70!black}{(+6.9)}} &
        30.7 {\scriptsize\textcolor{red!70!black}{(+8.2)}} &
        0.470 {\scriptsize\textcolor{red!70!black}{(+.070)}} &
        0.620 {\scriptsize\textcolor{red!70!black}{(-.063)}} &
        0.291 {\scriptsize\textcolor{red!70!black}{(+.045)}} &
        0.823 {\scriptsize\textcolor{red!70!black}{(-.022)}} &
        0.10 {\scriptsize\textcolor{red!70!black}{(+.00)}} &
        0.89 {\scriptsize\textcolor{green!70!black}{(-.03)}} \\
        \bottomrule
    \end{tabular}
    }
\end{table}

\subsection{Mechanism Analyses}
\label{sec:mechanisms}

\begin{figure}[t]
    \centering
    \includegraphics[width=0.95\linewidth]{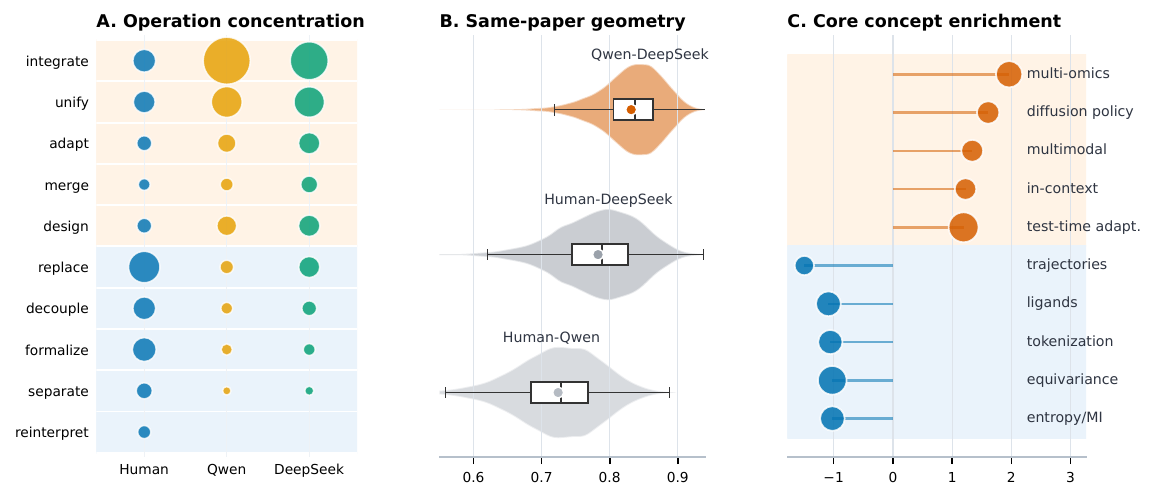}
    \caption{Mechanism analyses for the distributional gap. (A) Circle areas show archetype operation shares by source: model outputs concentrate on reusable template moves such as \textit{integrate} and \textit{unify}, while human ideas retain more local interventions such as \textit{replace}, \textit{decouple}, and \textit{formalize}. (B) Same-paper similarity distributions show that the two model ideas are closer to each other than either is to the human idea. (C) Core concept enrichment ranks clusters by model-vs-human log-odds, with circle area indicating support. Model-heavy concepts are high-frequency technical motifs, whereas human-heavy concepts are narrower mechanism or representation clusters.}
    \label{fig:mechanism_summary}
\end{figure}

The taxonomy tells us \textit{where} model and human distributions differ, but it does not fully explain \textit{why} LLMs repeatedly choose bridge-and-synthesis moves. We therefore run follow-up analyses on human ideas and two representative model sources, \texttt{Qwen3-8B} and \texttt{DeepSeek-V4-Flash}. \autoref{fig:mechanism_summary} suggests that the gap is not simply that models \textit{do more synthesis}. 
Instead, model ideas concentrate around a recognizable recipe: select a high-frequency technical concept cluster and apply a safe synthesis operation such as integrating, unifying, combining, or adapting it with another nearby concept.

\paragraph{Archetype Clustering.}
We use \texttt{GPT-5.4-mini} to rewrite each proposal into a one-sentence archetype that abstracts away domain-specific details while preserving the high-level idea. We then cluster these valid archetypes using TF-IDF and MiniBatchKMeans with \(k=30\)~\citep{sculley2010web}, and normalize the main verb in each archetype into an operation family. The strongest model-heavy operation is \textbf{\textit{integrate}}: it appears 7,994 times in model outputs (34.2\%) but only 275 times in human ideas (2.35\%), giving a model-vs-human log-odds of 3.07. Other overrepresented operations are also synthesis-like, including \textbf{\textit{unify}} (8.2\% vs. 1.9\%, log-odds 1.52), \textbf{\textit{design}} (1.5\% vs. 0.3\%, 1.50), \textbf{\textit{merge}} (1.37), and \textbf{\textit{adapt}} (1.36).

The human-heavy side looks different. Human ideas more often use local intervention operations such as \textbf{\textit{replace}}, \textbf{\textit{decouple}}, and \textbf{\textit{formalize}}. Specifically, \textbf{\textit{replace}} makes up 9.13\% of human operations versus just 0.92\% for models, and \textbf{\textit{decouple}} reaches 2.33\% among humans and only 0.21\% for models. Two human-majority clusters are largely missing from model outputs: \textbf{\textit{replace}} (83.3\% human) and \textbf{\textit{decouple}} (85.4\%). Relative to model-heavy clusters such as integrate / model / learning, they achieve higher bottleneck specificity (2.61, 2.70) and lower boilerplate (0.60, 0.51). Beyond reduced synthesis, human ideas typically revise targeted assumptions, modules and mechanisms in local research settings.

\paragraph{Representation Mechanism.}
We map proposals, motivations, methods, prior works, archetypes and extracted concept phrases into a shared representation space with 2560-dimensional embeddings with \texttt{Qwen3-Embedding-4B}. These representations show strong consensus between model outputs from the same source paper. For a given input paper, the cosine similarity between ideas from \texttt{Qwen3-8B} and \texttt{DeepSeek-V4-Flash} reaches 0.8316, while the scores stand at 0.7242 for human–Qwen pairs and 0.7829 for human–Deepseek pairs. The model pair exceeds the average human–model similarity, suggesting that distinct model families converge to similar generation patterns even when given the same context and paper-specific prior knowledge as humans.

We quantify how diffusely each proposal is positioned relative to its prior-work set. Given a proposal embedding $p$ and prior-work embeddings $\{w_i\}_{i=1}^K$, we compute cosine similarities $s_i=p^\top w_i$, convert them into a similarity distribution, and compute its normalized entropy $H$. Higher $H$ means that the proposal is comparably close to several prior works, rather than being dominated by a single reference. We also compute $B = p^\top c + H - (s_{(1)} - s_{(2)})$, where $c$ is the normalized centroid of the prior-work embeddings, and $s_{(1)}$ and $s_{(2)}$ are the largest and second-largest similarities $B$ increases when a proposal is close to the overall prior-work centroid, has high entropy over priors, and its top two related-work similarities are not sharply separated. Human proposals have higher values on both quantities ($H=0.7215$, $B=1.4662$) than model proposals, with \texttt{Qwen3-8B} at $H=0.6494$ and $B=1.3345$, and \texttt{DeepSeek-V4-Flash} at $H=0.6745$ and $B=1.4237$. This suggests that model proposals are less broadly positioned within the available prior-work set.

\paragraph{Concept Enrichment.}
To connect archetype operations with their semantic content, we rank the representation-analysis concept clusters by model-vs-human log-odds, keeping clusters with at least 30 occurrences. We report both \textit{core} archetype concepts and \textit{all} concepts, which add proposal-level TF-IDF terms. Since the pool contains one human and two model sources, we interpret enrichment by log-odds instead of raw majority.

The strongest model-enriched core clusters are reusable technical motifs: multi-omics integration has 317 occurrences across 180 records and 95.0\% model share (log-odds 1.96), followed by diffusion policy (93.1\%, 1.61), multimodal generation (91.2\%, 1.34), in-context learning (90.2\%, 1.23), test-time adaptation / adaptive optimization (89.5\%, 1.19), quantization, multi-agent / LLM-agent concepts, and multimodal reasoning (86.6--88.1\%). The all-concept view keeps multi-omics and diffusion policy at the top and adds physical constraints, climate models, adversarial robustness, implicit regularization, and cross-modal representations. Many representative phrases already contain \textit{integrate}, \textit{combine}, or \textit{unify}, suggesting that these concepts serve as slots for the synthesis template.

Human-enriched clusters are more local: trajectories and tracking trajectories (63.5\% human, log-odds -1.50), ligands and molecular interactions, tokenization and token importance, equivariance / inverse problems / Hamiltonian structure, entropy / mutual information, routing and prototypes, and verification concepts (44.8 to 53.2\%, -0.75 to -1.09). In the all-concept view, function vectors / internal representations, policy evaluation, denoising, and geometric concepts remain relatively human-enriched. These clusters name variables, constraints, and mechanisms to intervene on, together with the operation analysis, they suggest that LLMs start from high-frequency technical concepts and wrap them in safe integration moves, while human papers more often modify, separate, or formalize a narrower local mechanism.

The mechanism analyses suggest 
that the gap between LLMs and human is not random, nor simply a preference for legitimate interdisciplinary work. LLM ideas repeatedly instantiate an archetype-level recipe: choose a salient technical concept cluster, then integrate or unify it with another nearby object. Human ideas are more likely to make a narrower intervention, such as replacing a brittle component, decoupling two confounded mechanisms, or formalizing a local structure. 
This provides one plausible account of why LLM ideas can seem reasonable while still occupying a more concentrated region of research taste.

\section{Conclusion and Discussion}

We introduced an evaluation framework for comparing human and LLM research ideas under shared literature-grounded inputs. By extracting human ideas from real papers, prompting LLMs on reconstructed related-work contexts, and labeling ideas with a two-axis research-taste taxonomy, we find that LLMs occupy a much narrower region of research taste than humans. The central pattern is an overproduction of bridge-like opportunities and synthesis-oriented method paradigms, supported by diagnostic and mechanism analyses showing that many model outputs are less specific, more boilerplate-like, and organized around repeated integrate archetypes. LLM ideation should be evaluated as a distributional alignment problem. A useful ideation system should also diversify how it identifies problems, constructs contributions, and departs from familiar synthesis templates. This gives future ideation systems a target: preserve the fluency and scale of LLM generation while shifting proposals toward more specific, mechanism-aware, and less template-bound research patterns.

\section*{Limitations}

Our corpus is broad but still STEM-centered, so research-taste distributions may differ in social science, humanities, clinical research, or engineering design. Our task reconstructs a local literature context from related works, whereas real researchers draw on tacit expertise, failed attempts, collaborations, reviewer feedback, and long-term research agendas. Although our taxonomy and LLM annotation pipeline are human-validated, they compress nuanced ideas into discrete labels and diagnostic scores, and some proposals naturally mix multiple research moves. Finally, we evaluate a finite set of models, prompts, and one-shot settings; interactive agents, domain-specific systems, retrieval-heavy pipelines, or prompt interventions may reduce some of the observed gaps. Our results should therefore be read as evidence about current LLM ideation under a controlled literature-grounded setting, not as a claim about all possible AI-assisted research workflows.

\section*{Acknowledgments}
This work was supported in part by the U.S. National Science Foundation under award No. 2541654.

\bibliographystyle{unsrtnat}
\bibliography{main}

\newpage
\appendix

\section{Dataset Details}
\label{app:dataset}

Each data point starts from a real human paper and contains its extracted idea, represented as a motivation and method, together with a reconstructed local literature context. The literature context consists of proximal prior works, each represented by title and abstract. We collect papers from ICLR, ICML, NeurIPS, and Nature Communications, then merge the human source, all model generations, and all annotations by paper ID. Rows with missing or invalid model outputs or labels are removed. The final matched evaluation set contains 11,683 papers: 5,994 ML papers and 5,689 Nature Communications papers as shown in \autoref{tab:dataset_stats}.

\begin{table}[H]
    \centering
    \caption{Dataset statistics for the matched evaluation corpus. Papers are rows used in the main distributional analyses, where all evaluated sources have valid outputs and annotations. Avg. priors and median priors report the number of reconstructed prior works per paper.}
    \label{tab:dataset_stats}
    \begin{tabular}{lcccc}
        \toprule
        Corpus & Source years & Papers & Avg. priors & Median priors \\
        \midrule
        ML Conference & 2023--2026 & 5,994 & 6.52 & 7 \\
        Nature Communications & 2023--2025 & 5,689 & 5.89 & 6 \\
        Mixed & 2023--2026 & 11,683 & 6.21 & 6 \\
        \bottomrule
    \end{tabular}
\end{table}

\section{Taxonomy Design}
\label{app:taxonomy}

The taxonomy was designed to compare research taste rather than topic, field, or technical substrate. Before coding examples, we reviewed proposal-writing and research-gap guidance from DARPA's Heilmeier Catechism,\footnote{\url{https://www.darpa.mil/about/heilmeier-catechism}} NIH application guidance,\footnote{\url{https://www.grants.nih.gov/grants-process/write-application/advice-on-application-sections}} the NSF PAPPG project-description instructions,\footnote{\url{https://www.nsf.gov/policies/pappg/24-1/ch-2-proposal-preparation}} and AHRQ guidance on identifying research gaps.\footnote{\url{https://www.ncbi.nlm.nih.gov/books/NBK62480/}} These documents shaped the distinction between identifying why current knowledge is insufficient and explaining what new contribution will address that insufficiency. We then iteratively refined labels on sampled human and LLM ideas, removing categories tied to domain or method family.

\paragraph{Opportunity Pattern.}
\begin{itemize}[leftmargin=1.2em,itemsep=1pt,topsep=2pt]
    \item \textit{Puzzle / Contradiction}: a paradox, tradeoff, surprising failure, or conflicting evidence.
    \item \textit{Explanation Gap}: a missing causal, mechanistic, theoretical, or explanatory account.
    \item \textit{Scope Mismatch}: unrealistic assumptions, narrow regimes, unclear transferability, or boundary conditions.
    \item \textit{Evidence Gap}: missing ways to observe, measure, benchmark, audit, diagnose, validate, compare, or accumulate evidence.
    \item \textit{Bridge Opportunity}: disconnected literatures, theories, evidence streams, communities, or methods that could be connected.
    \item \textit{Failure / Risk Gap}: brittleness, unreliability, bias, uncertainty, safety/privacy/security risk, or reproducibility concerns.
    \item \textit{Resource Bottleneck}: cost, compute, time, data access, sample scarcity, experimental burden, deployment friction, usability, or scalability constraints.
\end{itemize}

\paragraph{Method Paradigm.}
\begin{itemize}[leftmargin=1.2em,itemsep=1pt,topsep=2pt]
    \item \textit{Synthesis / Unification}: bridges, integrates, reconciles, or unifies separate literatures, theories, evidence streams, mechanisms, or methods.
    \item \textit{Relax / Extend Scope}: makes prior work function under weaker assumptions, broader scope, new regimes, noisier conditions, or more realistic settings.
    \item \textit{Robustification}: reduces failures, brittleness, risk, uncertainty, bias, unreliability, or trustworthiness problems.
    \item \textit{Formal Derivation}: introduces a formal model, theorem, bound, objective, derivation, proof, ontology, taxonomy, conceptual distinction, or explanatory formulation.
    \item \textit{Empirical Mapping}: builds or applies systematic measurement, benchmarks, diagnostics, datasets, empirical maps, comparative studies, or pattern analyses.
    \item \textit{Artifact / System}: builds a concrete artifact, software system, platform, device, material, prototype, or deployment workflow as the central contribution.
    \item \textit{Optimization / Search}: uses optimization, search, screening, tuning, active/adaptive design, scaling, resource allocation, or efficiency strategies to discover or improve a solution.
\end{itemize}

\begin{table}[t]
\centering
\small
\caption{Guideline primitives used to ground the research-taste taxonomy. We report the related items extracted from each source and how they informed our taxonomy.}
\begin{tabular}{p{0.16\linewidth}p{0.38\linewidth}p{0.38\linewidth}}
\toprule
\textbf{Source} & \textbf{Extracted Information} & \textbf{For Our Taxonomy} \\
\midrule
DARPA Heilmeier Catechism
& Limits of current practice; novelty of the approach; risk; cost and timeline; mid-term and final checks for success.
& Separates the motivating limitation from the proposed approach, and directly motivates labels for scope limits, risk/failure, resource bottlenecks, and success-oriented evidence. \\
\midrule
NIH Application Guidance
& Specific Aims; hypothesis-based goals; scientific question; Significance; Innovation; Approach; major experiments; rigor, reproducibility, transparency; facilities and resources.
& Supports the motivation--method split through Significance/Innovation versus Approach, and motivates labels involving explanation, empirical work, robustness/reproducibility, and resources. \\
\midrule
NSF PAPPG Project Description
& Objectives and methods; potential to advance knowledge; relationship to the present state of knowledge; work plan and experimental procedures; how success will be known; research products such as data, software, models, samples, and equipment; interdisciplinary work and integration or transfer of knowledge; facilities and equipment.
& Gives the statement of the proposal pattern: what, why, how, success criteria, and benefits. It supports evidence, artifact/system, resource, bridge/synthesis, and scope-oriented categories. \\
\midrule
AHRQ Research-Gap Framework
& Why a research gap exists; where evidence falls short; insufficient or imprecise information; biased information; inconsistent or unknown consistency; not-the-right information; population, intervention, comparison, outcome, and setting elements; translation from gaps to research questions.
& Provides the grounding for the opportunity axis, especially evidence gaps, contradictions, scope mismatch, risk/bias, empirical characterization, and scope extension. \\
\bottomrule
\end{tabular}
\label{tab:taxonomy_guideline_grounding}
\end{table}

\section{Experimental Details and Configurations}
\label{app:responsible_use}

\paragraph{Artifacts and intended use.}
This work uses public scholarly artifacts: paper metadata, titles, abstracts, extracted prior-work contexts, model-generated ideas, taxonomy labels, and annotation outputs. Source papers are drawn from ML conference proceedings and Nature Communications; abstracts are retrieved from public scholarly metadata providers when available. We use these artifacts only for research evaluation. Released dataset will preserve source attribution.

\paragraph{Privacy and content.}
The corpus does not contain private user-generated text, recruited participant records, or human-subject data. It consists of public scholarly text and derived research-idea summaries. 

\paragraph{Model and compute setup.}
Each evaluated model generates one idea per input context. The main proposal-generation prompt uses the detailed profile in \autoref{fig:generation_prompt}. Local open-weight runs use temperature \(0.6\), top-\(p=0.95\), top-\(k=20\), maximum 2,048 new tokens, and one output per input; the GPT API run uses temperature \(1.0\) and a JSON-schema output constraint. Thinking-mode runs differ only in the model-side thinking flag or reasoning setting. API models are run through provider APIs; local model and embedding jobs are run on cluster GPUs. The embedding analysis uses \texttt{Qwen3-Embedding-4B} with 2,560-dimensional embeddings, maximum length 512, batch size 12, bfloat16, and last-token pooling on CUDA.

\paragraph{Analysis parameters.}
No hyperparameter search is performed for the reported distributional metrics. Archetype clustering uses TF-IDF with lowercasing, English stop words, 1--2 grams, \(\mathrm{min\_df}=2\), \(\mathrm{max\_df}=0.85\), sublinear TF, and MiniBatchKMeans with \(k=30\), batch size 512, seed 13, and \texttt{n\_init=auto}. Concept extraction uses TF-IDF over proposal text with 1--3 grams and custom stop words, while concept clustering uses MiniBatchKMeans over Qwen3 embeddings with batch size 512 and seed 13. Linear probes use 5-fold stratified cross-validation with balanced logistic regression, \(C=1.0\), \texttt{liblinear}, and fixed random seed 13.

\paragraph{Human audit instructions.}
Two authors audited a stratified sample of 150 annotations. For each item, they were shown the prior-work titles, motivation, method, LLM-assigned labels, diagnostic scores, and the taxonomy definitions in \autoref{app:taxonomy}. They independently judged whether the primary Opportunity Pattern, primary Method Paradigm, and diagnostic-score profile were acceptable under the codebook, then disagreements were inspected through confusion matrices. No external annotators were recruited or paid.

\paragraph{AI-assistant use.}
LLMs are part of the research pipeline: they assist with idea extraction, related-work reconstruction, proposal generation, annotation, and archetype rewriting. The manuscript reports these uses in the method and prompt appendices. We use LLMs for writing polishing and grammar checking.

\section{Prompts for Idea Generation and Annotation}
\label{app:prompts}

We show prompt templates used in the pipeline. Bracketed fields are filled separately for each paper.

\begin{headerbox}[Prompt: Prior Work Extraction]{neutralgray}{neutralgraybg}
    \small
    You are an expert AI research analyst. Given a paper, identify the \textbf{proximal prior works} that most directly shaped its core idea. Focus on papers the authors likely engaged with when forming the contribution, not on general background knowledge.

    \medskip
    \textbf{Step 1: Identify the core idea.}\enspace
    Briefly determine the paper's main innovation, the prior limitation or gap it responds to, and the specific insight that makes the contribution non-obvious.

    \medskip
    \textbf{Step 2: Select proximal prior works.}\enspace
    Choose \textbf{5--7} papers from the related-work context that are most useful for reconstructing how the current idea could have emerged. Prefer recent and specific predecessors over classic, foundational, or textbook-style works.

    \medskip
    For each candidate prior work, apply three checks:
    \begin{enumerate}[leftmargin=1.5em,itemsep=1pt,topsep=2pt]
        \item \textbf{Counterfactual check}: Would the authors likely still have produced this specific idea without reading this paper? If yes, exclude it.
        \item \textbf{Specificity check}: What concrete method, finding, limitation, dataset, or insight from this prior work informed the current paper?
        \item \textbf{Proximity check}: Is this the most recent or direct source for that influence? Prefer the newer and more proximal paper.
    \end{enumerate}

    \medskip
    \textbf{Do not include:}
    \begin{itemize}[leftmargin=1.5em,itemsep=0pt,topsep=2pt]
        \item classic foundational works cited only as background;
        \item generic tools, datasets, libraries, or infrastructure;
        \item papers used only as baselines or mentioned only in passing;
        \item papers that share the topic but do not shape the core idea.
    \end{itemize}

    \medskip
    \textbf{Output} (JSON): return 5--7 prior works. For each, provide \texttt{cite\_id}, title, authors, year, and a one-sentence explanation of what the authors learned from it and how it informed the current paper.
\end{headerbox}
\begin{center}
\begin{minipage}{0.98\linewidth}
\captionof{figure}{Prompt used to reconstruct the proximal prior works for each human paper.}
\label{fig:prior_work_prompt}
\end{minipage}
\end{center}

\begin{headerbox}[Prompt: LLM Idea Generation]{neutralgray}{neutralgraybg}
    \small
    You are a research scientist skilled at synthesizing ideas from existing literature into novel research proposals. You are given a set of related research papers with titles and abstracts. Analyze these papers, identify research gaps and opportunities, and propose one coherent novel research idea.

    \medskip
    \textbf{Input format.}\enspace
    The user message lists prior works as blocks of \texttt{\# Title (cite\_id)} followed by \texttt{\#\# Abstract: ...}.

    \medskip
    \textbf{Output requirements.}\enspace
    Based only on the papers above, return a valid JSON object with exactly two string fields:
    
    \begin{quote}
    \ttfamily\small
    \{\newline
    \hspace*{1em}"motivation": "...",\newline
    \hspace*{1em}"method": "..."\newline
    \}
    \end{quote}
    
    The motivation should synthesize the research gap, why it matters, and why the listed works leave room for the proposed idea. The method should describe a concrete, feasible high-level approach and explain how it addresses the gap. Do not include citations outside the provided papers, markdown fences, or any text before or after the JSON object.
\end{headerbox}
\begin{center}
\begin{minipage}{0.98\linewidth}
\captionof{figure}{Prompt template used to generate LLM research ideas from the reconstructed prior-work context.}
\label{fig:generation_prompt}
\end{minipage}
\end{center}

\begin{headerbox}[Prompt: Research-Taste Annotation]{neutralgray}{neutralgraybg}
    \small
    You are an expert annotator of research taste. Label the proposal using high-level categories that apply across ML/AI, natural science, medicine, engineering, and social or behavioral science. Do not classify by topic, domain, or technical substrate. Classify the proposal's \textbf{problem-finding pattern} and \textbf{idea-construction paradigm}. The two axes use disjoint labels; never copy a method-paradigm label into the opportunity axis, or vice versa.

    \medskip
    \textbf{Opportunity Pattern labels.}
    \begin{itemize}[leftmargin=1.5em,itemsep=0pt,topsep=2pt]
        \item \textbf{Puzzle / Contradiction}: the gap comes from a paradox, tradeoff, surprising failure, or conflicting evidence.
        \item \textbf{Explanation Gap}: the proposal asks why something works, fails, varies, or appears.
        \item \textbf{Scope Mismatch}: prior work relies on narrow, unrealistic, or poorly transferable assumptions.
        \item \textbf{Evidence Gap}: the field lacks measurement, benchmarking, auditing, diagnosis, validation, or comparison.
        \item \textbf{Bridge Opportunity}: disconnected literatures, methods, theories, or evidence streams need to be connected.
        \item \textbf{Failure / Risk Gap}: existing approaches raise reliability, robustness, safety, bias, uncertainty, or reproducibility concerns.
        \item \textbf{Resource Bottleneck}: progress is limited by cost, compute, data, time, deployment, usability, or scalability.
    \end{itemize}

    \medskip
    \textbf{Method Paradigm labels.}
    \begin{itemize}[leftmargin=1.5em,itemsep=0pt,topsep=2pt]
        \item \textbf{Synthesis / Unification}: integrates or reconciles separate literatures, mechanisms, theories, evidence streams, or methods.
        \item \textbf{Relax / Extend Scope}: makes prior work apply under broader, noisier, weaker-assumption, or more realistic settings.
        \item \textbf{Robustification}: reduces failure, brittleness, risk, uncertainty, bias, unreliability, or trustworthiness problems.
        \item \textbf{Formal Derivation}: introduces a theory, theorem, bound, objective, proof, taxonomy, or conceptual formulation.
        \item \textbf{Empirical Mapping}: builds or applies measurements, benchmarks, datasets, diagnostics, empirical maps, or comparative studies.
        \item \textbf{Artifact / System}: constructs a concrete tool, platform, software system, device, material, prototype, or workflow.
        \item \textbf{Optimization / Search}: uses optimization, search, tuning, screening, selection, allocation, scaling, or efficiency strategy.
    \end{itemize}

    \medskip
    \textbf{Input fields.}\enspace
    Paper ID; prior-work titles for context; proposal motivation; proposal method.

    \medskip
    \textbf{Decision guidance.}\enspace
    The opportunity axis asks how the gap is found. The method-paradigm axis asks what kind of research move constructs the paper. Use Synthesis / Unification only when bridging or reconciling separate lines of work is central, not merely when a method has multiple components. Use Empirical Mapping for estimating, auditing, diagnosing, quantifying, or characterizing a phenomenon. Use Artifact / System only when a concrete artifact, system, tool, platform, material, or prototype is the central deliverable. Use Optimization / Search when the central move is efficiency, scaling, search, tuning, selection, allocation, or resource-aware design.

    \medskip
    \textbf{Output} (JSON): return primary and secondary labels for each axis, confidence scores, diagnostic scores for \texttt{surface\_stitching}, \texttt{bottleneck\_specificity}, and \texttt{boilerplate}, and one concise rationale sentence.
    
    {%
    \renewenvironment{quote}
      {%
        \list{}{%
          \leftmargin=0pt
          \rightmargin=0pt
          \topsep=5pt
          \partopsep=0pt
          \parsep=0pt
          \itemsep=0pt
        }%
        \item\relax
      }
      {\endlist}
    \begin{quote}
    \ttfamily\small
    \{\newline
    \hspace*{1em}"labels": \{\newline
    \hspace*{2em}"opportunity\_pattern": \{"primary": "\textless one opportunity\_pattern label\textgreater", "secondary": "\textless one opportunity\_pattern label or none\textgreater"\},\newline
    \hspace*{2em}"research\_idea\_paradigm": \{"primary": "\textless one research\_idea\_paradigm label\textgreater", "secondary": "\textless one research\_idea\_paradigm label or none\textgreater"\}\newline
    \hspace*{1em}\},\newline
    \hspace*{1em}"confidence": \{"opportunity\_pattern": \textless 0.0-1.0\textgreater, "research\_idea\_paradigm": \textless 0.0-1.0\textgreater\},\newline
    \hspace*{1em}"diagnostics": \{\newline
    \hspace*{2em}"surface\_stitching": \textless true or false\textgreater,\newline
    \hspace*{2em}"surface\_stitching\_score": \textless integer 0-3, where 3 is clearly superficial A+B stitching\textgreater,\newline
    \hspace*{2em}"bottleneck\_specificity": \textless integer 0-3, where 3 identifies a precise bottleneck, mechanism, or limiting factor\textgreater,\newline
    \hspace*{2em}"boilerplate\_score": \textless integer 0-3, where 3 is highly generic or boilerplate\textgreater\newline
    \hspace*{1em}\},\newline
    \hspace*{1em}"rationale": "\textless one concise sentence\textgreater"\newline
    \}
    \end{quote}
    }
    
\end{headerbox}
\begin{center}
\begin{minipage}{0.98\linewidth}
\captionof{figure}{Annotation prompt. The implementation uses the same label set and maps the human-readable labels shown here to internal JSON keys.}
\label{fig:annotation_prompt}
\end{minipage}
\end{center}

\section{Additional Distributional Analyses}
\label{app:additional_distributional_analyses}

\subsection{Domain-Specific Results}
\label{app:domain_results}

We present the main distributional comparison separately for the Machine Learning corpus and the Nature Communications corpus in \autoref{tab:ml_only} and \autoref{tab:nc_only}. 

\begin{table}[H]
    \small
    \centering
    \setlength{\tabcolsep}{12pt}
    \caption{ML distributional distances against the human distribution. Ent. is normalized entropy. Header arrows indicate the direction closer to the human distribution.}
    \label{tab:ml_only}
    \begin{tabular}{lcccccc}
        \toprule
        & \multicolumn{3}{c}{\textbf{Opportunity Pattern}} & \multicolumn{3}{c}{\textbf{Method Paradigm}} \\
        \cmidrule(lr){2-4}\cmidrule(lr){5-7}
        Source & TVD \(\downarrow\) & JSD \(\downarrow\) & Ent. \(\uparrow\) & TVD \(\downarrow\) & JSD \(\downarrow\) & Ent. \(\uparrow\) \\
        \midrule
        \rowcolor{black!3}
        Human & --- & --- & 0.952 & --- & --- & 0.968 \\
        \midrule
        \texttt{Claude-Sonnet-4.6} & \cellcolor{blue!8}\textbf{0.447} & \cellcolor{blue!8}\textbf{0.172} & \cellcolor{blue!8}\textbf{0.699} & \cellcolor{blue!8}\textbf{0.288} & \cellcolor{blue!8}\textbf{0.105} & \cellcolor{blue!8}\textbf{0.896} \\
        \texttt{Gemini-3.1-Pro} & 0.448 & 0.188 & 0.676 & 0.350 & 0.156 & 0.835 \\
        \texttt{GPT-OSS-20B} & 0.643 & 0.338 & 0.438 & 0.532 & 0.254 & 0.683 \\
        \texttt{GPT-OSS-120B} & 0.683 & 0.379 & 0.383 & 0.523 & 0.266 & 0.679 \\
        \texttt{GPT-5.4-mini} & 0.579 & 0.276 & 0.525 & 0.397 & 0.155 & 0.819 \\
        \texttt{Qwen3-8B} & 0.598 & 0.302 & 0.505 & 0.533 & 0.301 & 0.641 \\
        \texttt{Qwen3-8B-Think} & \cellcolor{red!8}0.719 & \cellcolor{red!8}0.428 & \cellcolor{red!8}0.321 & \cellcolor{red!8}0.640 & \cellcolor{red!8}0.393 & \cellcolor{red!8}0.518 \\
        \texttt{Qwen3-32B} & 0.560 & 0.266 & 0.553 & 0.512 & 0.280 & 0.666 \\
        \texttt{DeepSeek-V4-Flash} & 0.522 & 0.231 & 0.605 & 0.326 & 0.135 & 0.855 \\
        \texttt{DeepSeek-V4-Flash-Think} & 0.593 & 0.291 & 0.511 & 0.402 & 0.194 & 0.785 \\
        \texttt{DeepSeek-V4-Pro} & 0.586 & 0.296 & 0.542 & 0.399 & 0.175 & 0.809 \\
        \bottomrule
    \end{tabular}
\end{table}

\begin{table}[H]
    \small
    \centering
    \setlength{\tabcolsep}{12pt}
    \caption{Nature Communications distributional distances against the human distribution. Ent. is normalized entropy. Header arrows indicate the direction closer to the human distribution.}
    \label{tab:nc_only}
    \begin{tabular}{lcccccc}
        \toprule
        & \multicolumn{3}{c}{\textbf{Opportunity Pattern}} & \multicolumn{3}{c}{\textbf{Method Paradigm}} \\
        \cmidrule(lr){2-4}\cmidrule(lr){5-7}
        Source & TVD \(\downarrow\) & JSD \(\downarrow\) & Ent. \(\uparrow\) & TVD \(\downarrow\) & JSD \(\downarrow\) & Ent. \(\uparrow\) \\
        \midrule
        \rowcolor{black!3}
        Human & --- & --- & 0.822 & --- & --- & 0.790 \\
        \midrule
        \texttt{Claude-Sonnet-4.6} & 0.250 & 0.093 & 0.670 & 0.177 & \cellcolor{blue!8}\textbf{0.045} & 0.698 \\
        \texttt{Gemini-3.1-Pro} & \cellcolor{blue!8}\textbf{0.243} & \cellcolor{blue!8}\textbf{0.079} & \cellcolor{blue!8}\textbf{0.728} & \cellcolor{blue!8}\textbf{0.140} & 0.047 & \cellcolor{blue!8}\textbf{0.741} \\
        \texttt{GPT-OSS-20B} & 0.298 & 0.124 & 0.591 & 0.232 & 0.096 & \cellcolor{red!8}0.565 \\
        \texttt{GPT-OSS-120B} & 0.351 & 0.159 & 0.585 & 0.252 & 0.094 & 0.608 \\
        \texttt{GPT-5.4-mini} & 0.441 & 0.216 & \cellcolor{red!8}0.528 & 0.278 & 0.098 & 0.701 \\
        \texttt{Qwen3-8B} & 0.250 & 0.085 & 0.617 & 0.221 & 0.100 & 0.635 \\
        \texttt{Qwen3-8B-Think} & \cellcolor{red!8}0.454 & \cellcolor{red!8}0.221 & 0.540 & \cellcolor{red!8}0.314 & \cellcolor{red!8}0.173 & 0.657 \\
        \texttt{Qwen3-32B} & 0.299 & 0.127 & 0.593 & 0.215 & 0.105 & 0.656 \\
        \texttt{DeepSeek-V4-Flash} & 0.272 & 0.111 & 0.641 & 0.181 & 0.054 & 0.652 \\
        \texttt{DeepSeek-V4-Flash-Think} & 0.340 & 0.144 & 0.617 & 0.184 & 0.066 & 0.680 \\
        \texttt{DeepSeek-V4-Pro} & 0.279 & 0.128 & 0.626 & 0.182 & 0.054 & 0.665 \\
        \bottomrule
    \end{tabular}
\end{table}

\begin{table}[H]
    \small
    \centering
    \caption{Domain-specific percentages for the two labels most emphasized in the main text: Bridge Opportunity on the opportunity axis and Synthesis / Unification on the method-paradigm axis. Header arrows mark the direction closer to the human distribution.}
    \label{tab:domain_bridge_synthesis}
    \begin{tabular}{lcccc}
        \toprule
        Source & ML Bridge \(\downarrow\) & ML Synthesis \(\downarrow\) & NC Bridge \(\downarrow\) & NC Synthesis \(\downarrow\) \\
        \midrule
        \rowcolor{black!3}
        Human & 14.0 & 6.6 & 10.2 & 3.4 \\
        \midrule
        \texttt{Claude-Sonnet-4.6} & 58.7 & 35.5 & 35.2 & 11.4 \\
        \texttt{Gemini-3.1-Pro} & 58.8 & 41.6 & 34.5 & 13.2 \\
        \texttt{GPT-OSS-20B} & 78.4 & 44.0 & 35.9 & 8.8 \\
        \texttt{GPT-OSS-120B} & 82.3 & 51.1 & 45.2 & 11.3 \\
        \texttt{GPT-5.4-mini} & 71.9 & 37.8 & 54.2 & 19.5 \\
        \texttt{Qwen3-8B} & 73.8 & 59.9 & 24.3 & 16.4 \\
        \texttt{Qwen3-8B-Think} & 85.9 & 70.7 & 55.5 & 32.8 \\
        \texttt{Qwen3-32B} & 70.0 & 57.1 & 36.9 & 19.2 \\
        \texttt{DeepSeek-V4-Flash} & 66.2 & 35.5 & 37.4 & 8.8 \\
        \texttt{DeepSeek-V4-Flash-Think} & 73.3 & 46.7 & 44.2 & 13.7 \\
        \texttt{DeepSeek-V4-Pro} & 71.5 & 46.5 & 38.1 & 11.1 \\
        \bottomrule
    \end{tabular}
\end{table}

\subsection{Full-Paper Context Ablation}
\label{app:full_context}

The main experiments represent each proximal prior work with its title and abstract. To test whether the bridge-and-synthesis pattern is an artifact of this compressed context, we run a full-paper context ablation on 1,000 inputs: 500 sampled from the Machine Learning corpus and 500 sampled from the Nature Communications corpus. For each related-work paper, the model first reads the full text and produces a compact summary with detailed motivation, method and insight. We then use these model-generated full-paper summaries in place of abstracts as the context for idea generation, keeping the downstream annotation pipeline unchanged.

\autoref{tab:full_context_ablation} compares the original abstract-context condition with the full-paper-summary condition on the same subset for \texttt{Qwen3-8B} and \texttt{DeepSeek-V4-Flash}. The richer context does not remove the bridge-heavy opportunity pattern. For \texttt{Qwen3-8B}, bridge opportunities increase from 456 to 551 cases; for \texttt{DeepSeek-V4-Flash}, they increase from 489 to 521 cases. The listed method-paradigm categories also do not show a compensating shift toward formal derivation, empirical mapping, artifact construction, or optimization. 

\begin{table}[H]
    \centering
    \footnotesize
    \setlength{\tabcolsep}{7pt}
    \caption{Full-paper context ablation on a 1,000 paper subset. Original uses title and abstract context; full context replaces abstracts with model-generated summaries of the full papers. All taxonomy labels are shown; parentheses show count changes from the original condition. Metric rows report TVD/JSD against the same human reference distribution as \autoref{tab:main_gap}; Ent. is normalized entropy.}
    \label{tab:full_context_ablation}
    \begin{adjustbox}{max width=0.92\textwidth}
    \begin{tabular}{lllll}
        \toprule
        & \multicolumn{2}{c}{\texttt{Qwen3-8B}} & \multicolumn{2}{c}{\texttt{DeepSeek-V4-Flash}} \\
        \cmidrule(lr){2-3}\cmidrule(lr){4-5}
        Label & Original & Full context & Original & Full context \\
        \midrule
        \rowcolor{black!3}
        \multicolumn{5}{l}{\textbf{Opportunity Pattern}} \\
        Puzzle / Contradiction & 8 & 5 {\scriptsize\textcolor{green!70!black}{(-3)}} & 39 & 36 {\scriptsize\textcolor{green!70!black}{(-3)}} \\
        Explanation Gap & 353 & 284 {\scriptsize\textcolor{green!70!black}{(-69)}} & 281 & 246 {\scriptsize\textcolor{green!70!black}{(-35)}} \\
        Scope Mismatch & 17 & 21 {\scriptsize\textcolor{red!70!black}{(+4)}} & 21 & 27 {\scriptsize\textcolor{red!70!black}{(+6)}} \\
        Evidence Gap & 43 & 41 {\scriptsize\textcolor{green!70!black}{(-2)}} & 58 & 55 {\scriptsize\textcolor{green!70!black}{(-3)}} \\
        Bridge Opportunity & 456 & 551 {\scriptsize\textcolor{red!70!black}{(+95)}} & 489 & 521 {\scriptsize\textcolor{red!70!black}{(+32)}} \\
        Failure / Risk Gap & 85 & 69 {\scriptsize\textcolor{green!70!black}{(-16)}} & 94 & 95 {\scriptsize\textcolor{red!70!black}{(+1)}} \\
        Resource Bottleneck & 38 & 29 {\scriptsize\textcolor{green!70!black}{(-9)}} & 18 & 20 {\scriptsize\textcolor{red!70!black}{(+2)}} \\
        \addlinespace[2pt]
        \rowcolor{black!3}
        \midrule
        \multicolumn{5}{l}{\textbf{Method Paradigm}} \\
        Synthesis / Unification & 389 & 451 {\scriptsize\textcolor{red!70!black}{(+62)}} & 233 & 260 {\scriptsize\textcolor{red!70!black}{(+27)}} \\
        Relax / Extend Scope & 24 & 26 {\scriptsize\textcolor{red!70!black}{(+2)}} & 60 & 47 {\scriptsize\textcolor{green!70!black}{(-13)}} \\
        Robustification & 36 & 32 {\scriptsize\textcolor{green!70!black}{(-4)}} & 82 & 95 {\scriptsize\textcolor{red!70!black}{(+13)}} \\
        Formal Derivation & 23 & 9 {\scriptsize\textcolor{green!70!black}{(-14)}} & 29 & 28 {\scriptsize\textcolor{green!70!black}{(-1)}} \\
        Empirical Mapping & 255 & 229 {\scriptsize\textcolor{green!70!black}{(-26)}} & 251 & 229 {\scriptsize\textcolor{green!70!black}{(-22)}} \\
        Artifact / System & 242 & 234 {\scriptsize\textcolor{green!70!black}{(-8)}} & 286 & 286 {\scriptsize\textcolor{gray!70!black}{(0)}} \\
        Optimization / Search & 31 & 19 {\scriptsize\textcolor{green!70!black}{(-12)}} & 59 & 55 {\scriptsize\textcolor{green!70!black}{(-4)}} \\
        \addlinespace[2pt]
        \midrule
        \rowcolor{black!3}
        \multicolumn{5}{l}{\textbf{Opportunity Pattern Metrics}} \\
        TVD \(\downarrow\) & 0.376 & 0.430 {\scriptsize\textcolor{red!70!black}{(+.054)}} & 0.368 & 0.400 {\scriptsize\textcolor{red!70!black}{(+.032)}} \\
        JSD \(\downarrow\) & 0.165 & 0.205 {\scriptsize\textcolor{red!70!black}{(+.040)}} & 0.152 & 0.160 {\scriptsize\textcolor{red!70!black}{(+.008)}} \\
        Ent. \(\uparrow\) & 0.669 & 0.623 {\scriptsize\textcolor{green!70!black}{(-.046)}} & 0.706 & 0.701 {\scriptsize\textcolor{green!70!black}{(-.005)}} \\
        \addlinespace[2pt]
        \rowcolor{black!3}
        \midrule
        \multicolumn{5}{l}{\textbf{Method Paradigm Metrics}} \\
        TVD \(\downarrow\) & 0.338 & 0.400 {\scriptsize\textcolor{red!70!black}{(+.062)}} & 0.213 & 0.236 {\scriptsize\textcolor{red!70!black}{(+.023)}} \\
        JSD \(\downarrow\) & 0.182 & 0.229 {\scriptsize\textcolor{red!70!black}{(+.047)}} & 0.079 & 0.093 {\scriptsize\textcolor{red!70!black}{(+.014)}} \\
        Ent. \(\uparrow\) & 0.752 & 0.699 {\scriptsize\textcolor{green!70!black}{(-.053)}} & 0.867 & 0.860 {\scriptsize\textcolor{green!70!black}{(-.007)}} \\
        \bottomrule
    \end{tabular}
    \end{adjustbox}
\end{table}

\subsection{Prompt Ablation}
\label{app:prompt_ablation}

The main generation prompt asks models to analyze a set of prior papers, identify research gaps and opportunities, and propose one coherent idea. This wording may itself encourage models to connect papers into a synthesis-style proposal. We therefore run a prompt ablation that relaxes this constraint. We use more neutral terms like \textit{generate} and \textit{describe} instead of expressions that might affect the paradigm of the model ideation.

\begin{headerbox}[Prompt: Relaxed LLM Idea Generation]{neutralgray}{neutralgraybg}
    \small
    You are a research scientist skilled at generating ideas from existing literature into novel research proposals. You are given a set of related research papers with titles and abstracts. Analyze these papers, identify research gaps and opportunities, and propose one coherent novel research idea.

    \medskip
    \textbf{Input format.}\enspace
    The user message lists prior works as blocks of \texttt{\# Title (cite\_id)} followed by \texttt{\#\# Abstract: ...}.

    \medskip
    \textbf{Output requirements.}\enspace
    Based only on the papers above, return a valid JSON object with exactly two string fields:
    \begin{quote}
    \ttfamily\small
    \{\newline
    \hspace*{1em}"motivation": "...",\newline
    \hspace*{1em}"method": "..."\newline
    \}
    \end{quote}
    The motivation should describe the research gap, why it matters, and why the listed works leave room for the proposed idea. The method should describe a concrete, feasible high-level approach and explain how it addresses the gap. Do not include citations outside the provided papers, markdown fences, or any text before or after the JSON object.
\end{headerbox}
\begin{center}
\begin{minipage}{0.98\linewidth}
\captionof{figure}{Relaxed generation prompt used for the prompt ablation.}
\label{fig:prompt_ablation_prompt}
\end{minipage}
\end{center}

\autoref{tab:prompt_ablation} reports the resulting count changes on the full matched evaluation set. The ablation changes the distribution, but it does not eliminate the main qualitative pattern. For \texttt{Qwen3-8B}, bridge opportunities decrease from 5,807 to 5,247 cases, while failure / risk gaps increase from 987 to 1,238 cases; nevertheless, bridge opportunities remain the largest listed opportunity category. For \texttt{DeepSeek-V4-Flash}, bridge opportunities increase from 6,094 to 6,368 cases under the relaxed prompt. On the method axis, the relaxed prompt moves some mass toward artifact / system and optimization / search categories, especially for \texttt{Qwen3-8B}, but the shifts are modest relative to the overall corpus size. These results suggest that prompt wording affects the exact label mix, while the tendency to organize generated ideas around bridge-like opportunities remains stable across prompt variants.

\begin{table}[H]
    \centering
    \footnotesize
    \setlength{\tabcolsep}{7pt}
    \caption{Prompt ablation on the full matched evaluation set of 11,683 papers. Original uses the main generation prompt in \autoref{fig:generation_prompt}; relaxed uses the prompt in \autoref{fig:prompt_ablation_prompt}. All taxonomy labels are shown; parentheses show count changes from the original condition. Metric rows report TVD/JSD against the human reference distribution; Ent. is normalized entropy.}
    \label{tab:prompt_ablation}
    \begin{adjustbox}{max width=0.92\textwidth}
    \begin{tabular}{lllll}
        \toprule
        & \multicolumn{2}{c}{\texttt{Qwen3-8B}} & \multicolumn{2}{c}{\texttt{DeepSeek-V4-Flash}} \\
        \cmidrule(lr){2-3}\cmidrule(lr){4-5}
        Label & Original & Relaxed & Original & Relaxed \\
        \midrule
        \rowcolor{black!3}
        \multicolumn{5}{l}{\textbf{Opportunity Pattern}} \\
        Puzzle / Contradiction & 121 & 86 {\scriptsize\textcolor{green!70!black}{(-35)}} & 381 & 348 {\scriptsize\textcolor{green!70!black}{(-33)}} \\
        Explanation Gap & 3,714 & 3,917 {\scriptsize\textcolor{red!70!black}{(+203)}} & 3,064 & 3,017 {\scriptsize\textcolor{green!70!black}{(-47)}} \\
        Scope Mismatch & 203 & 280 {\scriptsize\textcolor{red!70!black}{(+77)}} & 305 & 333 {\scriptsize\textcolor{red!70!black}{(+28)}} \\
        Evidence Gap & 440 & 482 {\scriptsize\textcolor{red!70!black}{(+42)}} & 550 & 478 {\scriptsize\textcolor{green!70!black}{(-72)}} \\
        Bridge Opportunity & 5,807 & 5,247 {\scriptsize\textcolor{green!70!black}{(-560)}} & 6,094 & 6,368 {\scriptsize\textcolor{red!70!black}{(+274)}} \\
        Failure / Risk Gap & 987 & 1,238 {\scriptsize\textcolor{red!70!black}{(+251)}} & 1,106 & 957 {\scriptsize\textcolor{green!70!black}{(-149)}} \\
        Resource Bottleneck & 411 & 433 {\scriptsize\textcolor{red!70!black}{(+22)}} & 183 & 182 {\scriptsize\textcolor{green!70!black}{(-1)}} \\
        \addlinespace[2pt]
        \rowcolor{black!3}
        \midrule
        \multicolumn{5}{l}{\textbf{Method Paradigm}} \\
        Synthesis / Unification & 4,523 & 3,938 {\scriptsize\textcolor{green!70!black}{(-585)}} & 2,632 & 2,634 {\scriptsize\textcolor{red!70!black}{(+2)}} \\
        Relax / Extend Scope & 336 & 459 {\scriptsize\textcolor{red!70!black}{(+123)}} & 781 & 760 {\scriptsize\textcolor{green!70!black}{(-21)}} \\
        Robustification & 413 & 531 {\scriptsize\textcolor{red!70!black}{(+118)}} & 975 & 929 {\scriptsize\textcolor{green!70!black}{(-46)}} \\
        Formal Derivation & 168 & 204 {\scriptsize\textcolor{red!70!black}{(+36)}} & 226 & 223 {\scriptsize\textcolor{green!70!black}{(-3)}} \\
        Empirical Mapping & 2,570 & 2,612 {\scriptsize\textcolor{red!70!black}{(+42)}} & 2,661 & 2,549 {\scriptsize\textcolor{green!70!black}{(-112)}} \\
        Artifact / System & 3,402 & 3,597 {\scriptsize\textcolor{red!70!black}{(+195)}} & 3,871 & 4,031 {\scriptsize\textcolor{red!70!black}{(+160)}} \\
        Optimization / Search & 271 & 342 {\scriptsize\textcolor{red!70!black}{(+71)}} & 537 & 557 {\scriptsize\textcolor{red!70!black}{(+20)}} \\
        \addlinespace[2pt]
        \midrule
        \rowcolor{black!3}
        \multicolumn{5}{l}{\textbf{Opportunity Pattern Metrics}} \\
        TVD \(\downarrow\) & 0.382 & 0.351 {\scriptsize\textcolor{green!70!black}{(-.031)}} & 0.400 & 0.424 {\scriptsize\textcolor{red!70!black}{(+.024)}} \\
        JSD \(\downarrow\) & 0.179 & 0.152 {\scriptsize\textcolor{green!70!black}{(-.027)}} & 0.167 & 0.181 {\scriptsize\textcolor{red!70!black}{(+.014)}} \\
        Ent. \(\uparrow\) & 0.658 & 0.690 {\scriptsize\textcolor{red!70!black}{(+.032)}} & 0.683 & 0.661 {\scriptsize\textcolor{green!70!black}{(-.022)}} \\
        \addlinespace[2pt]
        \rowcolor{black!3}
        \midrule
        \multicolumn{5}{l}{\textbf{Method Paradigm Metrics}} \\
        TVD \(\downarrow\) & 0.368 & 0.335 {\scriptsize\textcolor{green!70!black}{(-.033)}} & 0.246 & 0.260 {\scriptsize\textcolor{red!70!black}{(+.014)}} \\
        JSD \(\downarrow\) & 0.190 & 0.153 {\scriptsize\textcolor{green!70!black}{(-.037)}} & 0.086 & 0.089 {\scriptsize\textcolor{red!70!black}{(+.003)}} \\
        Ent. \(\uparrow\) & 0.734 & 0.774 {\scriptsize\textcolor{red!70!black}{(+.040)}} & 0.845 & 0.840 {\scriptsize\textcolor{green!70!black}{(-.005)}} \\
        \bottomrule
    \end{tabular}
    \end{adjustbox}
\end{table}

\end{document}